\documentclass[sigconf]{acmart}
\AtBeginDocument{%
  }
\usepackage{acmart-taps}
\setcopyright{acmlicensed}
\copyrightyear{2024}
\acmYear{2024}
\setcopyright{acmlicensed}\acmConference[SIGGRAPH Conference Papers '24]{Special Interest Group on Computer Graphics and Interactive Techniques Conference Conference Papers '24}{July 27-August 1, 2024}{Denver, CO, USA}
\acmBooktitle{Special Interest Group on Computer Graphics and Interactive Techniques Conference Conference Papers '24 (SIGGRAPH Conference Papers '24), July 27-August 1, 2024, Denver, CO, USA}
\acmDOI{10.1145/3641519.3657455}
\acmISBN{979-8-4007-0525-0/24/07}

\usepackage{float}
\usepackage{wrapfig}
\usepackage{enumitem}
\usepackage{bm}
\usepackage{multirow}
\usepackage{makecell}
\usepackage{pifont}

\newcommand{\delete}[1]{}
\newcommand{\addmine}[1]{#1}

\begin{document}

\title{RTG-SLAM: Real-time 3D Reconstruction at Scale Using Gaussian Splatting}

\author{Zhexi Peng}
\authornote{Joint first authors}
\orcid{0000-0003-4342-5263}
\email{zhexipeng@zju.edu.cn}
\affiliation{
\institution{State Key Lab of CAD\&CG\\Zhejiang University}
\city{Hangzhou}
\country{China}
}

\author{Tianjia Shao}
\authornotemark[1]

\orcid{0000-0001-5485-3752}
\email{tjshao@zju.edu.cn}
\affiliation{
\institution{State Key Lab of CAD\&CG\\Zhejiang University}
\city{Hangzhou}
\country{China}
}

\author{Yong Liu}
\orcid{0009-0000-8133-5686}
\author{Jingke Zhou}
\orcid{0009-0004-0754-573X}
\email{zilae@zju.edu.cn}
\email{zhoujk@zju.edu.cn}
\affiliation{
\institution{State Key Lab of CAD\&CG\\Zhejiang University}
\city{Hangzhou}
\country{China}
}



\author{Yin Yang}
\email{yin.yang@utah.edu}
\orcid{0000-0001-7645-5931}
\affiliation{
\institution{University of Utah}
\city{Salt Lake City}
\country{USA}
}

\author{Jingdong Wang}
\email{welleast@gmail.com}
\orcid{0000-0002-4888-4445}
\affiliation{
\institution{Baidu Research}
\city{Beijing}
\country{China}
}

\author{Kun Zhou}
\authornote{Corresponding author}
\email{kunzhou@acm.org}
\orcid{0000-0003-4243-6112}
\affiliation{
\institution{State Key Lab of CAD\&CG\\Zhejiang University}
\city{Hangzhou}
\country{China}
}

\renewcommand{\shortauthors}{Peng et al.}

\begin{abstract}
  We present Real-time Gaussian SLAM (RTG-SLAM), a real-time 3D reconstruction system with an RGBD camera for large-scale environments using Gaussian splatting. The system features a compact Gaussian representation and a highly efficient on-the-fly Gaussian optimization scheme. We force each Gaussian to be either opaque or nearly transparent, with the opaque ones fitting the surface and dominant colors, and transparent ones fitting residual colors. 
  By rendering depth in a different way from color rendering, we let a single opaque Gaussian well fit a local surface region without the need of multiple overlapping Gaussians, hence largely reducing the memory and computation cost. For on-the-fly Gaussian optimization, we explicitly add Gaussians for three types of pixels per frame: newly observed, with large color errors, and with large depth errors. We also categorize all Gaussians into stable and unstable ones, where the stable Gaussians are expected to well fit previously observed RGBD images and otherwise unstable. We only optimize the unstable Gaussians and only render the pixels occupied by unstable Gaussians. In this way, both the number of Gaussians to be optimized and pixels to be rendered are largely reduced, and the optimization can be done in real time. We show real-time reconstructions of a variety of large scenes. 
  Compared with the state-of-the-art NeRF-based RGBD SLAM, our system achieves comparable high-quality reconstruction but with around twice the speed and half the memory cost, and shows superior performance in the realism of novel view synthesis and camera tracking accuracy. 
\end{abstract}



\begin{teaserfigure}
  \center
  \includegraphics[width = \textwidth]{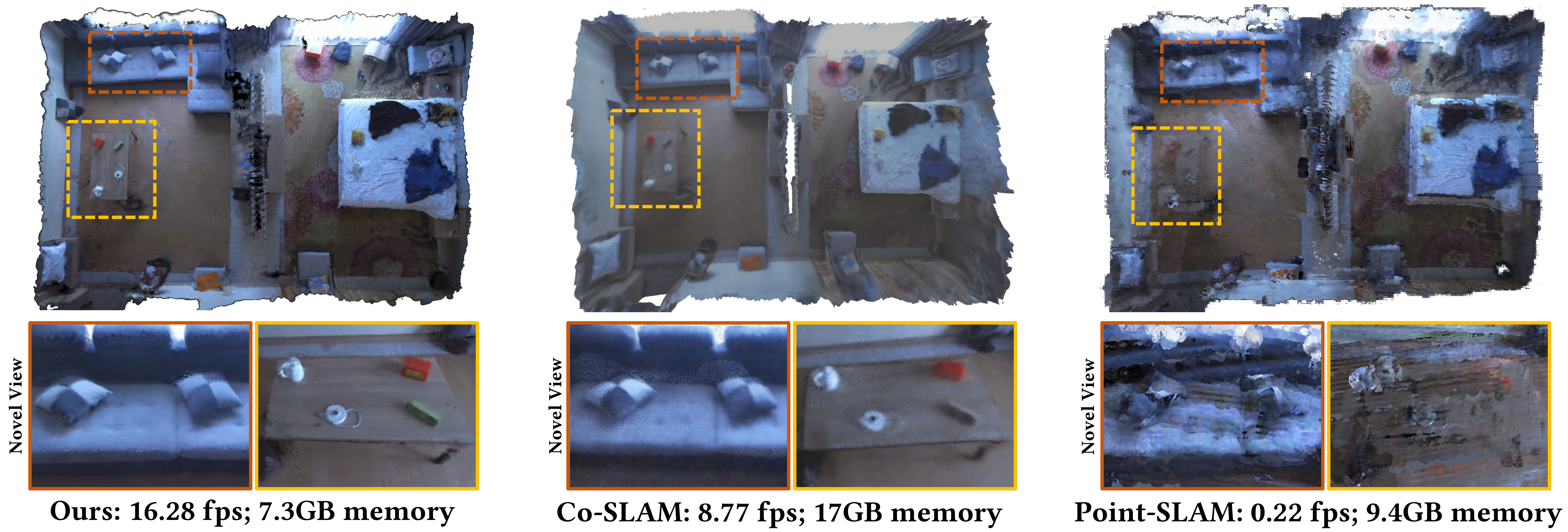}
  \Description{Figure 1. Fully described in the following caption.}
  \caption{
  A hotel room (about $56.3m^2$$\times$$1.7m$) reconstructed by our system and the state-of-the-art NeRF-based RGBD SLAM techniques (Co-SLAM~\cite{co_slam}, Point-SLAM~\cite{point_slam}) without any post-processing. Compared with the state-of-the-art NeRF-based RGBD SLAM, our system achieves comparable high-quality reconstruction but with around twice the speed and half the memory cost, and shows higher realism in novel view synthesis.  
  }
  \label{fig:teaser}
\end{teaserfigure}


\maketitle

\section{Introduction}\label{sec:intro}
Real-time 3D reconstruction at scale has been a long-studied problem in computer graphics and vision, and is crucial in many applications including VR/AR, autonomous robots, and interactive scanning with immediate feedback. 
With the ubiquity of RGBD cameras (e.g., Microsoft Kinect), different RGBD SLAM (Simultaneous Localization and Mapping) methods are proposed for real-time 3D reconstruction, using a variety of surface representations such as point clouds~\cite{DuHRCGSF11}, surfels~\cite{elasticfusion,pointfusion}, and signed distance functions~\cite{kinectfusion}. These methods are able to reconstruct large-scale scenes in real time with high-quality 3D surfaces~\cite{slam-octree,vox_hashing,bundlefusion}. However, they mainly focus on the geometry accuracy of the 3D reconstruction, and rarely consider the rendering realism of reconstructed results.

Some works attempt to employ neural radiance fields (NeRF) as the implicit scene representation~\cite{NeRF20} for dense RGBD SLAM in hopes of achieving high-quality reconstruction of both geometry and appearance. These methods typically represent the scene as an MLP network~\cite{DBLP:conf/iccv/SucarLOD21} or an implicit grid~\cite{Zhu2022CVPR, vox_fusion}, optimizing the scene parameters and estimating the camera pose via differentiable volume rendering. However, due to the expensive cost of volume rendering, these methods have difficulties in reaching real-time performance. Besides, the high memory cost makes it hard for them to handle large scale scenes. 

More recently, 3D Gaussians~\cite{3DGS} have emerged as an alternative representation of radiance fields, which can achieve equal or better rendering quality than previous NeRFs while being much faster in rendering and training. However, the 3D Gaussian representation up until now is mainly used in offline reconstruction scenarios~\cite{yang2023deformable3dgs,chung2024depthregularized}, not suitable for online reconstruction tasks with sequential RGBD inputs. 
To use it for real-time 3D reconstruction at scale, the core problem lies in how to represent the scene with low memory and computation cost, and how to perform online Gaussian optimization in real time.
We noticed there are several concurrent works~\cite{splatam,gs_dense_slam, gaussianslam,matsuki2023gaussian,photo_slam} trying to incorporate Gaussians into RGBD SLAM systems, which use different scene representations with Gaussians as well as different online optimization strategies. While promising results are demonstrated, there is still a long way to realize real-time reconstruction of large-scale scenes.

In this paper, we introduce Real-time Gaussian SLAM (RTG-SLAM), a real-time 3D reconstruction system with an RGBD camera for large-scale environments using Gaussian splatting, featuring a compact Gaussian representation and a highly efficient on-the-fly Gaussian optimization scheme.
In our compact Gaussian representation, we force each Gaussian to be either opaque or nearly transparent, with the opaque ones fitting the surface (i.e., depth map) and dominant colors, and transparent ones fitting residual colors. Our intention is to use a single opaque Gaussian to fit a local region of the surface without the need for multiple overlapping Gaussians. However, even for an opaque Gaussian, rendering its depth in the same way as rendering color would produce varying depth values declined from the Gaussian center, making it inaccurate to represent a local area using this Gaussian alone. To this end, we propose to render depth in a different way from color rendering. Following classical point rendering techniques~\cite{surfacesplatting}, we treat each opaque Gaussian as an ellipsoid disc on the dominant plane of Gaussian, so that it can well fit a local region or a large flat area by itself. The depth rendering is very convenient under this setting. During color rendering, we already have the sorted Gaussians as well as their opacities for each pixel. 
By selecting from front to back the first Gaussian whose opacity for the pixel is larger than a given threshold, we consider the ray hits the ellipsoid disc and compute the intersection point using equations of the ray and disc plane. Then the depth for the pixel is equal to the depth of the intersection point. The whole process is differentiable, so Gaussians can be optimized by measuring the differences between the rendered and input depth maps through backpropagation. 
The compact Gaussian representation can fit the 3D surfaces with much fewer Gaussians, hence largely reducing the memory and computation cost. 

We design a highly efficient on-the-fly Gaussian optimization scheme for the compact Gaussian representation.
We first categorize all Gaussians into stable and unstable ones following classical point-based reconstruction works~\cite{pointfusion}, based on whether they have been sufficiently optimized. The stable Gaussians are expected to well fit previously observed RGBD images and otherwise unstable.
Then given a new RGBD frame during scanning, instead of adaptively densifying Gaussians based on view space position gradients~\cite{3DGS}, we explicitly add Gaussians for three types of pixels with valid depths: newly observed pixels, pixels with large color errors after color re-rendering, and pixels with large depth errors after depth re-rendering.
For newly observed pixels or pixels with large depth errors, which means new opaque Gausians are required to fit the surface, we uniformly sample a small portion of pixels to initialize opaque Gaussians.
For the pixels with only large color errors, which means they already have opaque Gaussians well fitting the surface but poorly fitting the appearance in the current view, we apply the same pixel sampling and check the states of associated opaque Gaussians. If unstable, we leave them to continue being optimized. Otherwise, we add a transparent Gaussian to provide a residual color to improve the color in the current view without breaking previous observation. 
Afterwards, we launch the optimization process based on the re-rendering losses of color and depth. Note we only optimize the unstable Gaussians and only render the pixels occupied by the unstable Gaussians. In this way, both the number of Gaussians to be optimized and pixels to be rendered are largely reduced, and the optimization can be done in real time. We also establish a state management mechanism that enables the mutual conversion between stable/unstable Gaussians, as well as the removal of long-term erroneous Gaussians. Finally, to achieve accurate tracking in complex real-world environment, we use the classical frame-to-model ICP as the front-end odometry, and maintain a set of landmarks for back-end graph optimization.

We show real-time reconstructions of a variety of real large scenes, including corridor, storeroom, hotel room, home and office, ranging from $43 m^2$$\sim$$100 m^2$. All the results are scanned and reconstructed with a Microsoft Azure Kinect in real time (around 16 fps) without any post-processing. 
Comparisons demonstrate that RTG-SLAM runs at around twice the speed of the state-of-the-art NeRF-based SLAM, with around half the memory cost (e.g., 17.9 fps, 8.8 GB versus 8.65 fps, 17.3 GB~\cite{co_slam} on the home scene).
We also compare our method with the concurrent Gaussian SLAM work SplaTAM~\cite{splatam} (the only one with code published). We also surpass SplaTAM in speed and memory, where SplaTAM runs at 0.31 fps and is out of memory during scanning of the home scene.
We also conduct extensive experiments on three widely-used datasets: Replica~\cite{replica19arxiv}, TUM-RGBD~\cite{TUM} and ScanNet++~\cite{yeshwanthliu2023scannetpp}. 
Compared with the state-of-the-art NeRF SLAM methods, our system achieves comparable high-quality reconstruction, and shows superior performance in time and memory performance, realism of novel view synthesis, and camera tracking accuracy.

\section{Related Work}\label{sec:related}
\paragraph{Classical RGBD dense SLAM} 
There has been extensive work on 3D reconstruction with RGBD cameras over the past decade. We point the reader to the excellent state-of-the-art report~\cite{survey18} for detailed reviews. For online 3D reconstruction of scenes, numerous valuable works have emerged in the field of RGBD dense SLAM, with a variety of map representations, such as point clouds~\cite{DuHRCGSF11}, Hermite radial basis functions~\cite{HRBFFusion}, surfels~\cite{pointfusion,elasticfusion,Real-Time-High-Accuracy}, and signed distance functions (TSDF)~\cite{kinectfusion, vox_hashing, bundlefusion,octree13,ZhangXTZ15,SemanticPriors}. For example, BundleFusion~\cite{bundlefusion}, the state-of-the-art TSDF method for online reconstructing large-scale scenes, presents real-time globally consistent 3D reconstruction using on-the-fly surface re-integration, which reconstructs high-quality 3D scenes at scale. ElasticFusion~\cite{elasticfusion} represents scenes as a collection of surfels, employing surfel-rendered depth and color for tracking, also achieving high-quality results in real time. DI-Fusion~\cite{huang2021difusion} encodes scene priors considering both the local geometry and uncertainty parameterized by a deep neural network. These works mainly focus on the geometry reconstruction, while differently, our method simultaneously considers the surface reconstruction and photorealistic rendering.

\paragraph{NeRF-based RGBD dense SLAM}
Recently, with the great success of neural radiance fields (NeRF)~\cite{NeRF20}, some works have integrated NeRF with RGBD dense SLAM systems. 
For example, iMap~\cite{imap} is the first NeRF SLAM method using a single MLP as the scene representation. 
NICE-SLAM~\cite{nice_slam} represents scenes as hierarchical feature grids, utilizing pre-trained MLPs for decoding. 
Vox-fusion~\cite{vox_fusion} represents scenes as voxel-based neural implicit surfaces and stores them using octrees. The state-of-the-art NeRF SLAM works include ESLAM~\cite{eslam} representing scenes as multi-resolution feature grids, and Co-SLAM~\cite{co_slam} representing scenes as multi-resolution hash grids. 
An alternative approach is Point-SLAM~\cite{point_slam}, which employs neural point clouds and performs volumetric rendering with feature interpolation. These methods have achieved impressive results. However, as they are based on time-consuming volume rendering, all these methods have difficulties to reach real-time performance on real scenes. Besides, the memory cost of these NeRF SLAM methods is high, prohibiting them from reconstructing large-scale scenes.
In contrast, our method can reconstruct large scenes in real time, with much higher speed and lower memory cost.

\paragraph{Gaussian-based RGBD dense SLAM}
There are some concurrent works aiming to integrate 3D Gaussians into dense RGBD SLAM. 3D Gaussians~\cite{3DGS} can render high-quality images in real time, but the optimization is conducted offline typically requiring several minutes. To extend Gaussians to online reconstruction, \cite{gs_dense_slam} proposes an adaptive expansion strategy to add new
or delete noisy 3D Gaussian and a coarse-to-fine technique to select reliable Gaussians for tracking. \cite{gaussianslam} proposes novel strategies for seeding and optimizing Gaussian splats to extend their use to sequential RGBD inputs. SplaTAM~\cite{splatam} tailors an online reconstruction pipeline to use an underlying Gaussian representation and silhouette-guided optimization via differentiable rendering. \cite{matsuki2023gaussian} unifies the Gaussian representation for accurate, efficient tracking, mapping, and high-quality rendering. \cite{photo_slam} introduces a Gaussian-Pyramid-based training method to progressively learn multi-level features and enhance mapping performance.
While promising results are demonstrated, it is still difficult in reaching real-time reconstruction at scale. The reported fastest reconstruction speed is 8.34 fps on an NVIDIA RTX 4090 GPU~\cite{matsuki2023gaussian} on the synthetic Replica dataset~\cite{replica19arxiv}. They did not present the complete reconstruction results on real large scenes either. Thanks to our compact Gaussian representation and highly efficient Gaussian optimization strategy, our method can reconstruct real large scenes in real time with low memory cost.
\begin{figure*}[ht]
    \centering
    \includegraphics[width= \textwidth]{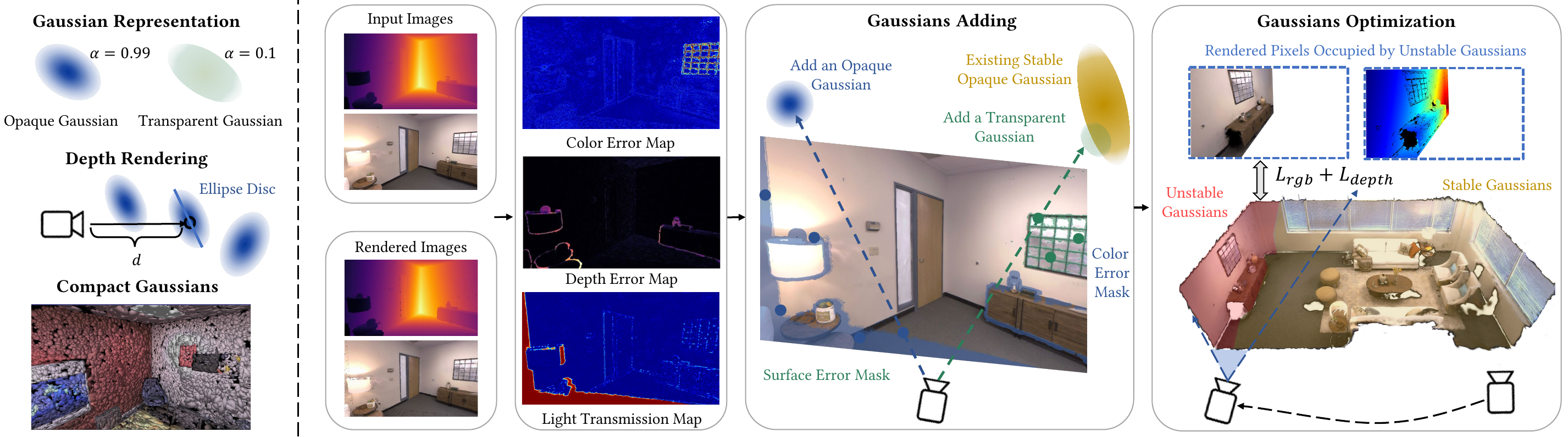}
    \Description{Figure 2. The overview of our method. Please refer to the following caption and text for details}
    \caption{Overview of our method. Left: we force each Gaussian to be either opaque or nearly transparent, and the depth is rendered differently from the color using the opaque Gaussian, so that a single opaque Gaussian can well fit a local region of the surface, yielding a compact Gaussian representation fitting 3D surfaces with much fewer Gaussians. Right: we compute the color error map, depth error map, and light transmission map to determine where to add opaque Gaussians or transparent Gaussians. we only optimize the unstable Gaussians, and only render the pixels occupied by them for optimization.}
    \label{fig:pipeline}
\end{figure*}

\section{Method}\label{sec:method}

The overview of our reconstruction pipeline is illustrated in Fig.~\ref{fig:pipeline}. In Sec. \ref{Map_Representation}, we first introduce our compact Gaussian representation and the corresponding rendering process of color and depth (Fig.~\ref{fig:pipeline} left). Next, we describe in detail the entire online reconstruction process based on the compact Gaussian representation in Sec. \ref{SLAM_process}.

\subsection{Compact Gaussian Representation}\label{Map_Representation}
We represent the scene $\mathcal{S}$ using a collection of 3D Gaussians $\{G_i\}$. Similar to \cite{3DGS}, each Gaussian is associated with the position $\mathbf{p}_i$, covariance matrix $\boldsymbol {\Sigma}_i$, opacity $\alpha_i$ and spherical harmonics (SH) coefficients $\mathbf{SH}_i$. The covariance matrix $\boldsymbol {\Sigma}_i$ is decomposed into a scale vector $\mathbf{s}_i$ and a quaternion $\mathbf{q}_i$. Each Gaussian is determined once after being added to be opaque ($\alpha=0.99$) for fitting the 3D surface and dominant color, or to be nearly transparent ($\alpha=0.1$) for fitting the residual color. 

We also treat each Gaussian as an ellipsoid disc (or surfel), and record the surfel parameters including  the normal $\mathbf{n}_i$, the confidence count $\eta_i$, and the initialization timestamp $t_i$. The normal vector is defined as the direction of the smallest eigenvector. The shape of surfel is defined as the region with Gaussian density larger than $\delta_{\alpha} = e^{-0.5}$ on the dominant plane of Gaussian, which corresponds to the density range within the standard deviation of the Gaussian distribution.
$\eta$ records how often a Gaussian is optimized, and $t$ records the time a Gaussian is created. We also divide the Gaussians into stable ones $\mathcal{S}_{stable}$ and unstable ones $\mathcal{S}_{unstable}$ based on the confidence count threshold $\delta_{\eta}$. All parameters are stored in a flat array indexed by the Gaussian index $i$. 

\paragraph{Image rendering}
The core of optimizing Gaussians lies in rendering color and depth maps through differentiable splatting, calculating errors with input RGBD images, and updating the Gaussian parameters. Now we introduce the rendering process in detail. Given a camera pose $\mathbf{T}_g$ and camera intrinsic matrix $\mathbf{K}$, the ray through the center of each pixel $\mathbf{u}$ in the image is defined as:
\begin{equation*}
\mathbf{r}(\mathbf{u}) = (\mathbf{R}_{g} \mathbf{K}^{-1} \dot{\mathbf{u}})\theta +\mathbf{t}_{g}, \; \text{where}\, \mathbf{T}_{g} = 
\left[
\begin{array}{cc}
     \mathbf{R}_{g} &  \mathbf{t}_{g}\\
     0 & 1 
\end{array}
\right]\in\mathbb{SE}(3).
\end{equation*}
Here $\theta$ is the length parameter along the ray direction and $\dot{\mathbf{u}}$ is the homogeneous vector $\dot{\mathbf{u}} := (\mathbf{u}^\top|1)^\top$. Then the color image $\hat{\mathbf{C}}$ can be rendered by alpha-blending proposed in \cite{3DGS} :
\begin{equation}
\begin{split}
\hat{\mathbf{C}}(\mathbf{u})=\sum_{i=1}^n \mathbf{c}_i f_i(\mathbf{u}) \prod_{j=1}^{i-1}\left(1-f_j(\mathbf{u})\right),
\end{split}
\end{equation}
where $\mathbf{c}_i$ represents the Gaussian color based on the view direction $\mathbf{r}_i$ and the SH coefficients $\mathbf{SH}_i$. $f_i(x)$ is computed by the center $\bm{\mu}_i$ and covariance matrix $\bm{\Sigma}_{2D, i}$ of the splatted 2D Gaussian in pixel space:
\begin{equation}
f(\mathbf{u})=\alpha_i \exp (-\frac{1}{2}(\mathbf{u}-\bm{\mu})^\top\bm{\Sigma}_{2D, i}^{-1}(\mathbf{u}-\bm{\mu})) \label{color_f}.
\end{equation}
Also a light transmission image $\hat{\mathbf{T}}$ to determine the visibility can be rendered as:
\begin{equation}
 \hat{\mathbf{T}}(\mathbf{u})= \prod_{i=1}^{n}\left(1-f_i(\mathbf{u})\right).
\end{equation}
$\hat{\mathbf{T}}$ represents the remaining energy of the light after it passes through a series of 3D Gaussians.

\begin{figure}[t]
    \centering
\includegraphics[width=0.9\columnwidth]{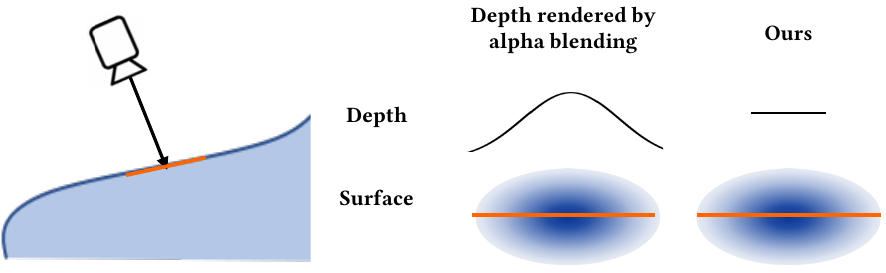}
    \Description{Figure 3. An illustration of our depth rendering method and it is described in detail in the following caption}
    \caption{If the depth is rendered in the same way as the color, the opaque Gaussian would produce varying depth values declined from the Gaussian center, making it inaccurate to represent a local surface. In contrast, we treat the opaque Gaussian as an ellipsoid disc on the dominant plane, and can well fit the local region.
    }
    \label{fig:depth_render}
\end{figure}

The depth rendering is the key for our compact Gaussian representation, where each single Gaussian can well fit a local region of surface without the need for multiple Gaussians. Note that all concurrent Gaussian SLAM works utilize the alpha blending methods to render the depth as the color. However, as illustrated in Fig.~\ref{fig:depth_render}, in the alpha blending setting, a single Gaussian will present varying depth values declined from the
Gaussian center, which is inappropriate to alone fit a local area that can typically be approximated as a plane. To this end, we render depth differently from rendering color. That is, for each pixel, we compute the intersection point of the view ray and the frontest opaque ellipsoid disc to obtain the pixel's depth.
Fortunately, we don't need to explicitly convert the Gaussians into ellipsoid discs and compute the intersections for each ray. During color rendering, all Gaussians $\{G^\mathbf{r}_j\}$ crossed by the ray $\mathbf{r}(\mathbf{u})$ are already sorted from front to back and the corresponding opacities $\{\alpha^\mathbf{r}_j\}$ along the ray are computed. The intersected Gaussian $G^\mathbf{r}_j$ is the first Gaussian with $\alpha^\mathbf{r}_j > \delta_{\alpha}$. The intersection point $\mathbf{p}_{G_j^\mathbf{r},\mathbf{r}}$ can be easily calculated by the ray plane intersection formula:
\begin{equation}
\mathbf{p}_{G_j^\mathbf{r},\mathbf{r}} = (\mathbf{R}_{g} \mathbf{K}^{-1} \dot{\mathbf{u}})\theta_{\mathbf{u}} +\mathbf{t}_{g}, \; \text{where}\, \theta_{\mathbf{u}}=\frac{(\mathbf{p}_j^\mathbf{r}-\mathbf{t}_{g})\cdot \mathbf{n}_j^\mathbf{r}}{(\mathbf{R}_{g} \mathbf{K}^{-1}\dot{\mathbf{u}})\cdot \mathbf{n}_j^\mathbf{r}}.
\end{equation}
Here $\mathbf{p}_j^\mathbf{r}$ and $\mathbf{n}_j^\mathbf{r}$ are the position and normal of intersected Gaussian.

If all $\{\alpha^\mathbf{r}_j\}$ are smaller than $\delta_{\alpha}$, the pixel depth is set to -1. When the disc normal and the ray are nearly perpendicular, the ray plane intersection could lead to aberrations and we use $\mathbf{p}_j^\mathbf{r}$ to approximate the intersection. Finally, depth map $\hat{\mathbf{D}}$ is defined as:
\begin{equation}\label{depth_map_rendering_eq}
\hat{\mathbf{D}}(\mathbf{u})=
\begin{cases}
         -1 \quad  \text{if no intersection}, \\
         (\mathbf{T}^{-1}_g \mathbf{p}_{G_j^\mathbf{r},\mathbf{r}})_z \quad  \text{elif}\; \langle\mathbf{n}_j^\mathbf{r}, \mathbf{r}\rangle < 60^{\circ},\\
         (\mathbf{T}^{-1}_g \mathbf{p}_j^\mathbf{r})_z \quad \text{otherwise}.
\end{cases}
\end{equation}

Clearly, the depth rendered by the Eq.~\eqref{depth_map_rendering_eq} is linearly computed from the position and rotation of the Gaussian, and such calculation process is fully differentiable, so the Gaussians can be optimized from the depth image loss. 
In addition, based on the results of ray-ellipsoid intersection, we also obtain the normal map $\hat{\mathbf{N}}(\mathbf{u})$ and index map $\hat{\mathbf{I}}(\mathbf{u})$, where the respective Gaussians' normals and indices are stored. $\hat{\mathbf{N}}(\mathbf{u})$ is employed to carry out the frame-to-model ICP for camera tracking, while the $\hat{\mathbf{I}}(\mathbf{u})$ provides a one-to-one mapping from pixels to Gaussians, utilized in subsequent processes such as Gaussian adding and state management.

\subsection{Online Reconstruction Process}\label{SLAM_process}
As shown in Fig. \ref{fig:pipeline}, our online reconstruction system is composed of the following stages.
\paragraph{Input pre-processing} Given an input color image $\mathbf{C}_k$ and depth map $\mathbf{D}_k$, following \cite{kinectfusion}, we compute the local vertex map $\mathbf{V}_k^l$ and the local normal map $\mathbf{N}_k^l$. With the estimated camera pose $\mathbf{T}_{g,k}$, $\mathbf{V}_k^l$ and $\mathbf{N}_k^l$ can be transformed into $\mathbf{V}_k^g$ and $\mathbf{N}_k^g$ in the global coordinate.

\paragraph{Gaussians adding} In order to obtain a complete representation of the environment, we need to add new Gaussians to the scene during online scanning to cover new observed regions. The adaptive control of Gaussians in \cite{3DGS} based on view-space positional gradients are inefficient for the online scanning. Therefore, we utilize a more efficient and reliable Gaussian adding strategy based on both geometry and appearance. Specifically, given the estimated camera pose $\mathbf{T}_{g,k}$, we first render the color map $\hat{\mathbf{C}}_k$, depth map $\hat{\mathbf{D}}_k$, light transmission map $\hat{\mathbf{T}}_k$ and index map $\hat{\mathbf{I}}_k$ using the existing Gaussians in the scene.
Then a mask $M$ is created to determine for which pixel a Gaussian should be added:
\begin{align}
    &M_{s} =\{\mathbf{u}_s\big|\hat{\mathbf{T}}_k(\mathbf{u}_{s}) > \delta_\mathbf{T}, \,\text{or } \,|\hat{\mathbf{D}}_k (\mathbf{u}_{s})- \mathbf{D}_k(\mathbf{u}_s)| > \delta_d\}, \nonumber\\ 
    &M_{c} = \{\mathbf{u}_c\big||\hat{\mathbf{C}}_k(\mathbf{u})- \mathbf{C}_k(\mathbf{u})| > \delta_\mathbf{c}, \,\text{and }\, \mathbf{u}_c\notin M_{s} \}.
\end{align}
Here $M_{s}$ represents regions where new geometry should be added.
$\hat{\mathbf{T}}_k(\mathbf{u}_{s}) > \delta_\mathbf{T}$ means the remaining energy of the ray is large without hitting any Gaussian or just hitting the Gaussian boundary, indicating newly observed areas, and $\delta_\mathbf{T} = 0.5$ in our setting.
$|\hat{\mathbf{D}}_k (\mathbf{u}_{s})- \mathbf{D}_k(\mathbf{u}_{s})| > \delta_d$ means there exist large re-rendering errors of depth, indicating new surface appears different from the existing scene, and 
$\delta_d = 0.1$ in our setting.
$M_{c}$ represents those areas that are geometrically accurate but with apparent color errors, and $\delta_c = 0.1$. 

With $\mathbf{V}_k^g$ and $M$, we have a good estimation of where those Gaussians should be added. However, adding Gaussians using all $M$ will cause considerable GPU memory overhead and hinder real-time performance. Therefore, we uniformly sample 5\% pixels on $M_{s}$ and $M_{c}$ to perform Gaussian adding. 
For the pixels sampled from $M_{s}$, we add Gaussians for them to fit the newly observed surface and we set the opacity to 0.99. For each pixel sampled from $M_{c}$, it is already associated with an existing opaque Gaussian in the scene, and we query that Gaussian using the index map $\hat{\mathbf{I}}_k$ and check its confidence state. If unstable, which means the Gaussian can be further optimized to fit the color, we do not add a new Gaussian for this pixel. Otherwise, we add a transparent ($\alpha=0.1$) Gaussian to correct color errors together with the stable Gaussian. The advantage of using transparent Gaussians lies in that such Gaussians with low opacity do not cause a significant attenuation of light energy, and the color impact to other views is little. In addition, during depth rendering, they are automatically filtered out by $\delta_{\alpha}$, therefore not affecting the depth rendering. All the added Gaussians are initialized as thin circle discs with the pixels' colors, positions, and normals, with confidence count $\eta=0$ and timestamp $t=k$.
For opaque Gaussians, their sizes are initialized to cover the scene as much as possible with little overlapping, while the transparent one's radius is limited to below 0.01\emph{m} to elliminate the disruption of the rendering results on other pixels. 

\paragraph{Gaussian optimization}
After adding Gaussians for the frames within a time window, we launch the Gaussian optimization based on the color loss and depth loss between the input and rendered RGBD images. We randomly sample a frame $k$ within the window per iteration during optimization. We use the $L_1$ loss for optimization:
\begin{equation}
    L_{color} = |\mathbf{C}_k - \hat{\mathbf{C}}_k|, \quad L_{depth} = |\mathbf{D}_k - \hat{\mathbf{D}}_k|.
\end{equation}
 The opacity learning rate $lr_{\alpha}$ is set to 0 to fix opacity and we do not calculate depth loss on pixels with no intersection with Gaussians. At the same time, we hope the transparent Gaussians focus on refining the local color without affecting other areas. Hence, we design a regularization term $L_{reg}$, an $L_2$ loss applied to all transparent Gaussians to constrain their geometry properties $\mathbf{p}, \mathbf{q}, \mathbf{s}$ remaining the same as their initial values.
The overall loss function is defined as:
\begin{equation} \label{total_loss}
    L=w_cL_{color} + w_dL_{depth} + w_{reg}L_{reg}.
\end{equation}

We use $w_c=1, w_d=1, w_{reg} = 1000$ in all our tests. We use the regularization term instead of zero learning rate for transparent Gaussians simply because 
in our PyTorch implementation it is difficult to set different learning rates for different Gaussians. The Gaussians will be optimized through multiple iterations and the confidence count is incremented by 1 when $\mathbf{SH}$ is updated. We notice that after optimization, the Gaussians can fit the current time window well but the rendering quality of previous views will decline, making it challenging to obtain high realism rendering under all views. Therefore we use a weighted average method to fuse the current result with previous results. 
Denote each Gaussian after the optimization for the current window $o$ as $G_o^{\prime}$,
and the fused Gaussian properties are computed as:
\begin{align}
    &G_o =(1-w_{curr})G_{o-1} + w_{curr} G^{\prime}_o, \nonumber \\
    &\eta_{o} = \eta^{\prime}_o, w_{curr} = \frac{\eta^{\prime}_o - \eta_{o-1}}{\eta^{\prime}_o}.
\end{align}
The fusing strategy can effectively avoid the forgetting problem. Note optimization of all the Gaussians is still too time-consuming for real-time reconstruction. To this end, we consider Gaussians with $\eta_k > \delta_{\eta}$ stable and otherwise unstable. The stable Gaussians have well fit previous observations and will not be optimized, greatly reducing the number of Gaussians that need to be optimized. Meanwhile, we only need to focus on the pixels affected by the unstable Gaussians, avoiding the optimization on all pixels. In this way, the optimization can be done in real time. Please refer to the supplementary material for more details.

\paragraph{State management}
Another key step is the mutual conversion between $\mathcal{S}_{stable}$ and $\mathcal{S}_{unstable}$ and the deletion of wrong Gaussians. We use the optimized scene $\mathcal{S}^*$ to render the color image $\hat{\mathbf{C}}_k^*$, depth map $\hat{\mathbf{D}}_k^*$, normal map $\hat{\mathbf{N}}_k^*$, and index map $\hat{\mathbf{I}}_k^*$ for frame $k$. We also calculate the $L_1$ difference for the color and depth compared with the RGBD input. For each stable Gaussian, if the corresponding color or depth error exceeds $\delta_\mathbf{c}$ or $\delta_d$, the error count $e_i$ of this Gaussian is incremented by 1. 
Then we manage the Gaussian states according to the following conditions.
The stable Gaussians with $e_i>\delta_{e}$ are converted to unstable while the unstable Gaussians with $\eta_i>\delta_{\eta}$ are converted to stable.
The unstable Gaussians with $k - t_i > \delta_{t}$ are removed because they keep unstable for a long time, and are treated as outliers. Note the Gaussians have to be observed in certain views to be marked as stable. Even if they are occluded later all the time, they are not redundant.

\paragraph{Camera tracking}
We utilize the frame-to-model ICP as the front-end odometry for camera tracking. Specifically, we use the optimized Gaussians in the previous frame to render the depth map $\hat{\mathbf{D}}_{k-1}^*$ and normal map $\hat{\mathbf{N}}_{k-1}^*$, and convert $\hat{\mathbf{D}}_{k-1}^*$ to the global space $\hat{\mathbf{V}}_{k-1}^{g*}$. Then given the current frame $\mathbf{V}_{k}^l$, we aim to find the camera pose that minimizes the point-to-plane error between 3D back-projected vertices:
\begin{equation}\label{eq:icp}
E(\boldsymbol{\xi}) = \sum \left\| \big(\mathbf{T}_{g, k} \mathbf{V}_k^l(\mathbf{u}) -\hat{\mathbf{V}}_{k-1}^{g*}(\hat{\mathbf{u}})\big) \cdot \hat{\mathbf{N}}_{k-1}^*(\hat{\mathbf{u}})\right\|.
\end{equation}
Here $\boldsymbol{\xi}$ is the Lie algebra representation of $\mathbf{T}_{g, k}$. We run a multi-level ICP to solve the objective function as \cite{kinectfusion}. Meanwhile, in order to reduce the drift during the scanning of large scenes, we also run a back-end optimization thread similar to ORB-SLAM2~\shortcite{ORBSLAM2}. While the pose estimation is finished, a set of 3D landmarks are also maintained. These landmarks are used for graph optimization in the back-end, enabling more accurate camera tracking.
\paragraph{Keyframes and global optimization}
In the global optimization, we further optimize the Gaussians in the global scene. Our keyframe selection strategy is inspired by \cite{Real-Time-High-Accuracy}. The keyframe list is constructed based on the camera motion. If the rotation angle relative to the last keyframe exceeds a threshold $\delta_{angle}$, or the relative translation exceeds $\delta_{move}$, we add a new keyframe. We use $\delta_{angle}=30^\circ$ and $\delta_{move}=0.3$\emph{m} in our experiments. Whenever a new keyframe is added, we optimize all the Gaussians in $\mathcal{S}$ using the latest keyframe and three randomly selected keyframes based on the same loss function as \eqref{total_loss}. In order to ensure the speed of global optimization, we only optimize the pixels with the top $40\%$ color errors on each keyframe. What's more, to avoid overfitting in the selected viewpoint, we do not update the position of the Gaussian during the global optimization and we use 0.1$\times$ the original learning rate to optimize the other parameters. When the scan is finished, we optimize $\mathcal{S}$ using all recorded keyframes with  $10\times$ the number of keyframes iterations.

\section{Evaluation}\label{sec:result}
\subsection{Experimental Setup}
\paragraph{Implementation Details} We implemented our SLAM system on a desktop computer with an \texttt{intel i9 13900KF} CPU and an \texttt{Nvidia RTX 4090} GPU. We implemented the mapping and tracking parts in \texttt{Python} using \texttt{Pytorch} framework and wrote custom \texttt{CUDA} kernels for rasterization and back propagation. We used an Azure Kinect as the RGBD camera for real-time scanning. Please refer to the supplementary material for more details.

\paragraph{Datasets} We conducted experiments on three public datasets: Replica~\cite{replica19arxiv}, TUM-RGBD~\cite{TUM}, ScanNet++~\cite{yeshwanthliu2023scannetpp}, and a self-scanned Azure dataset. Replica is the simplest benchmark as it contains synthetic, highly accurate and complete RGBD images. 
TUM-RGBD is a widely used dataset in the SLAM field for evaluating tracking accuracy because it provides accurate camera poses from an external motion capture system. 

ScanNet++ is a large-scale dataset that couples together capture of high-quality and commodity-level geometry and color of indoor scenes. Its depth maps are rendered from models reconstructed from laser scanning. Different from other benchmarks, each camera pose in ScanNet++ is very far apart from one another. 
We also scanned real-world scenes by ourselves to build an Azure dataset, including corridor, storeroom, hotel room, home, office, ranging from $43m^2$$\sim$$100m^2$. 

\paragraph{Baselines} We compare our method with existing state-of-the-art NeRF RGBD SLAM methods such as NICE-SLAM~\cite{nice_slam}, Point-SLAM~\cite{point_slam}, Co-SLAM~\cite{co_slam}, ESLAM~\cite{eslam} and a concurrent Gaussian SLAM work SplaTAM~\cite{splatam} (the only one with code released). We reproduce the results using the official code and run all experiments on the same desktop computer. Most of the experimental parameters follow their settings on the Replica dataset and we only adjust the bounding box setting based on the size of the new scenes. For ScanNet++, we double the scene (or map in SLAM) update frequency for all methods to ensure a fair comparison because of the sparsity of viewpoints.
\subsection{Evaluation of Online Reconstruction}
\paragraph{Time/memory performance}
Following \cite{point_slam}, we report the time per iteration for mapping optimization (e.g., NeRF optimization and Gaussian optimization), the tracking and mapping time per frame, the whole reconstruction FPS, the maximum memory usage during the SLAM process, and the final size of reconstructed scene on Replica office 0 and the home scene (around 70$m^2$) of our Azure dataset in Table \ref{time_and_memory}. 

We can see our reconstruction speed is around twice that of NeRF SLAM methods and about $46\times$ that of SplaTAM which is also based on 3D Gaussians. Notably, the memory cost of our method is much smaller compared to other methods, which allows us to scan large-scale environments. Note SplaTAM uses alpha blending to render depths as colors, thus yielding much more Gaussians (7155880 before out of memory in the home scene) than our compact Gaussian representation (987524). Even though they store the RGB values instead of spherical harmonics to reduce the memory overhead, their memory cost is still very high and runs out of memory in the home scene.
{\small 
    \begin{table}
            \caption{Comparison of time and memory performance on Replica (Off 0) and Azure Dataset (Home). Here \ding{53} means out of memory.}
        \centering
        \tabcolsep=0.15cm
        \resizebox{\columnwidth}{!}{
        \renewcommand\arraystretch{1.2}
        \begin{tabular}{c l c c c c c c c }
        \toprule[1pt]
            Method                     & Dataset & \makecell{Tracking\\/Frame} & \makecell{Mapping\\/Iteration} & \makecell{Mapping\\/Frame} & \makecell{FPS}   & \makecell{Model\\Size (MB)}& \makecell{Memory\\Cost (MB)}  \\ \hline
            \multirow{2}{*}{\makecell{NICE-SLAM\\\shortcite{nice_slam}}} & Replica & 1.05s            &  60.9ms         & 1.03s     &  0.95   &   87         &  9890          \\
                                       & Azure   & 0.68s            &  116.5ms        & 1.58s     &  0.63   &  \underline{136}         & 10057            \\\hline
            \multirow{2}{*}{\makecell{Co-SLAM\\\shortcite{co_slam}}}   & Replica & \underline{0.11s}            &  \underline{7.8ms}          & \underline{0.10s}      &  \underline{9.26}     &   \textbf{7}          &  \underline{7899}          \\
                                       & Azure   & \underline{0.11s}            &  \underline{7.2ms}          & 0.12s      &  \underline{8.65}     &   \textbf{7}         &  17342 \\\hline
            \multirow{2}{*}{\makecell{ESLAM\\\shortcite{eslam}}}     & Replica & 0.15s            &  16.7ms         & \underline{0.10s}     &  6.80     &   \underline{46}         &  18777          \\
                                       & Azure   & 0.13s            &  15.4ms         & \underline{0.11s}      &  7.54      &   139                                                          &  \ding{53} \\\hline
            \multirow{2}{*}{\makecell{Point-SLAM\\\shortcite{point_slam}}}& Replica & 1.05s           &  38.1ms        & 2.27s      &  0.44     &  15431     &  9890           \\
                                       & Azure   & 4.54s            &  68.4ms        & 4.00s      &  0.22     &  42536     &  \underline{9950}         
    \\\hline
            \multirow{2}{*}{\makecell{SplaTAM\\\shortcite{splatam}}}   & Replica &   1.16s         &   32.1ms     &  1.96s    &   0.51    &    310            &  9166           \\
                                       & Azure   &   2.00s        &   53.4ms     &   3.22s    &  0.31     &   520             &   \ding{53}  \\\hline
            \multirow{2}{*}{Ours}      & Replica & \textbf{0.02s}              &  \textbf{3.5ms}    & \textbf{0.05s}      &  \textbf{17.24}   &   71              &  \textbf{2751}            \\
                                       & Azure   & \textbf{0.03s}                       &  \textbf{4.3ms}                & \textbf{0.05s}     &  \textbf{17.90}            &   399             & \textbf{8782}            \\

        \bottomrule[1pt]
        \end{tabular}
        }
        \label{time_and_memory}
    \end{table} 
}

\paragraph{Tracking accuracy} The camera tracking accuracy on the real-world dataset TUM-RGBD is reported in Table \ref{track_acc_TUM}, and we report the results on the synthetic Replica dataset in the supplementary material.
Our method outperforms the NeRF SLAM methods and concurrent Gaussian SLAM method on both datasets, and achieves tracking accuracy comparable with classical SLAM methods on the real-world data. 

{\small 
    \begin{table}
        \caption{Comparison of tracking accuracy (unit: $cm$) on TUM-RGBD.}
        \centering
        \tabcolsep=0.15cm
        \renewcommand\arraystretch{1.2}
        \resizebox{0.95\columnwidth}{!}{
        \begin{tabular}{c c c c c}
        \toprule[1pt]
            Method     & fr1\_desk & fr2\_xyz & fr3\_office& Avg.   \\ \hline
            NICE-SLAM\shortcite{nice_slam}  & 4.30      & 31.73    &   3.87    & 13.28  \\
            Co-SLAM\shortcite{co_slam}    & 2.92      & 1.75     &   3.55    & 2.74   \\
            ESLAM\shortcite{eslam}      & 2.49      & 1.11     &   2.74    & 2.11   \\
            Point-SLAM\shortcite{point_slam} & 2.56      & 1.20     &   3.37    & 2.38   \\
            SplaTAM\shortcite{splatam}    & 3.33      & 1.55     &   5.28    & 3.39   \\
            Ours       & \underline{1.66}     & \textbf{0.38}      &  \underline{1.13}    & \underline{1.06}  \\ \hline
            ElasticFusion\shortcite{elasticfusion} & 2.53   & 1.17      &   2.52    & 2.07   \\
            ORB-SLAM2\shortcite{ORBSLAM2}  & \textbf{1.60}      &  \underline{0.40}      &   \textbf{1.00}    &  \textbf{1.00}   \\
            BAD-SLAM\shortcite{bad-slam}   & 1.70      & 1.10      &   1.70    & 1.50   \\

        \bottomrule[1pt]
        \end{tabular}
        }
        \label{track_acc_TUM}
    \end{table} 
}
\paragraph{Novel view synthesis} 

We qualitatively compare the rendering quality for novel view synthesis for all methods. The results are shown in Fig.~\ref{fig:scannetpp_compare}. Note the NeRF-based methods require a depth map to synthesize high-quality images, so we use the reconstructed mesh to render depth maps for them. We also quantitatively compare the novel-view synthesis on the ScanNet++ testing views, where the ground truth depth is used for NeRF-based methods, and the results are reported in the supplementary material. We can see that our method and SplaTAM clearly produces higher quality images with much fewer artifacts and higher fidelity appearance. 
We also quantitatively compare the rendering quality with other methods on the training views of Replica, following Point-SLAM and all concurrent Gaussian SLAM works. Please see the table in the supplementary material. Our method achieves a rendering quality comparable with SplaTAM and Point-SLAM (which needs the ground-truth depth map as input), and consistently outperforms the other NeRF SLAM methods.

\paragraph{Reconstruction quality} Following NICE-SLAM~\cite{nice_slam}, we use the metrics including \emph{Accuracy}, \emph{Completion}, \emph{Accuracy Ratio}[<3cm] and \emph{Completion Ratio}[<3cm] to evaluate the scene geometry on ScanNet++. We remove unseen regions that are not inside any camera's frustum. For the NeRF SLAM methods, the meshes produced by marching cubes with 1$cm$ voxel size are used for evaluation. For Point-SLAM, as mentioned in the paper, we use the re-rendered depth maps for TSDF Fusion. For SplaTAM and ours, we uniformly sample an equal amount of points from the reconstructed Gaussians for evaluation. To eliminate the impact of tracking accuracy, we use the ground truth camera pose for reconstruction and the results are reported in Table \ref{geometry_accuracy_scannet++}. Please note that Point-SLAM does not optimize the locations of neural points, so in this experiment its depth is always correct, thus always obtaining accurate geometry.
Nevertheless, our geometry accuracy outperforms other methods except Point-SLAM, and achieves comparable completion results. It demonstrates that our compact Gaussians can accurately fit surfaces with a small number of Gaussians.
We also demonstrate a qualitative comparison of reconstruction results and novel view synthesis in Fig.\ref{fig:ours_dataset_compare}. Note SplaTAM and ESLAM run out of memory in the scene. The top-view scenes in the teaser and Fig. \ref{fig:ours_dataset_compare} are directly rendered from Gaussians without mesh extraction for SplaTAM and ours.
We can see our method can achieve comparable high-quality reconstruction as the state-of-the-art NeRF SLAM methods, and surpass them in novel view synthesis. We further illustrate our reconstruction and novel view synthesis results on our real captured scenes in Fig.~\ref{fig:ours_more_res}.
{\small 
    \begin{table}
        \caption{Comparison of geometry accuracy on ScanNet++.} 
        \centering
        \tabcolsep=0.15cm
        \renewcommand\arraystretch{1.2}
        \resizebox{\columnwidth}{!}{

        \begin{tabular}{c c c c c }
        \toprule[1pt]
            Method          & Acc.$\downarrow$    & Acc. Ratio$\uparrow$   & Comp.$\downarrow$     & Comp. Ratio$\uparrow$ \\ \hline
            {NICE-SLAM\shortcite{nice_slam}}  & 4.45      & 74.49       & 2.04       &  86.63    \\ 
            {Co-SLAM\shortcite{co_slam}}   & 5.26      & 78.86       & 1.06       &  96.25    \\ 
            {ESLAM\shortcite{eslam}}     & 4.43      & 74.51       & \underline{1.05}       &  \underline{97.42}    \\ 
            {Point-SLAM\shortcite{point_slam}}   & \textbf{0.67}      & \textbf{99.12}       & \textbf{0.68}       &  \textbf{98.94}   \\ 
            {SplaTAM\shortcite{splatam}}     & 1.32      & 95.31       & 1.54       &  93.55    \\ 
            {Ours}       & \underline{0.95}      & \underline{96.41}       & 1.11       &  97.16     \\
        \bottomrule[1pt]
        \end{tabular}
        }
        \label{geometry_accuracy_scannet++}
    \end{table} 
}

\subsection{Ablation studies}
We evaluate the compact Gaussian representation here. Please see the supplementary material for the evaluation on stable/unstable Gaussians, and the sampled pixel number for Gaussians.
\paragraph{Compact Gaussian Representation} To prove the effectiveness of our compact Gaussian representation, we randomly select 20 RGBD images on Replica, and uniformly sample a certain number of pixels to initialize and optimize Guassians to fit the RGBD images. We compare the fitting results between our compact Gaussians and the original Gaussians using alpha blending. As shown in Fig.~\ref{fig:compact_ablation} left, our compact Gaussian representation requires much fewer Gaussians than the original Gaussian representation to reach the same depth accuracy. Also our compact Gaussian representation can better fit surfaces with the same amount of Gaussians (Fig.~\ref{fig:compact_ablation} right).

 We then assess the necessity of transparent Gaussians in the compact Gaussian representation. We show a reconstruction result versus the result trained using only opaque Gaussians in Figure \ref{fig:transparent_ablation}.
We can see that the pure opaque Gaussians will obscure the existing Gaussians during the color blending process, leading to significant color errors from new views.

\section{Conclusion}
We present a real-time 3D reconstruction system for large-scale environments using Gaussian splatting. We introduce a compact Gaussian representation to reduce the number of Gaussians needed to fit the surface, thereby greatly reducing the memory and computation cost. For on-the-fly Gaussian optimization, we explicitly add Gaussians for three types of pixels per frame: newly observed, with large color errors and with large depth errors, and only optimize the unstable Gaussians and only render the pixels occupied by unstable Gaussians. We reconstruct large-scale real scanning scenes, and achieve better performance than both the state-of-the-art NeRF SLAM method and the concurrent Gaussian SLAM methods. Because only opaque Gaussians and transparent Gaussians are used to represent the scene in order to reach real time reconstruction at scale, our rendering quality is inevitably degraded compared with original Gaussians. How to improve the rendering quality while keeping real-time performance is worth exploring in the future. Besides, the reflective or transparent materials can cause the surface color largely varying across different views, making some Gaussians frequently switch between two states and not optimized well. In the future we will also extend our system to handle outdoor scenes, dynamic objects, fast camera movement, and scenes with changing lightings.

\begin{acks}
The authors would like to thank the reviewers for their insightful comments. This work is supported by NSF China (No. U23A20311 \& 62322209), the XPLORER PRIZE, and the 100 Talents Program of Zhejiang University. The source code and data are available at \url{https://gapszju.github.io/RTG-SLAM}.
\end{acks}

\bibliographystyle{ACM-Reference-Format}
\bibliography{ref}
\clearpage

\aptLtoX[graphic=no,type=html]{\begin{figure*}
    \centering
    \includegraphics[width=\textwidth]{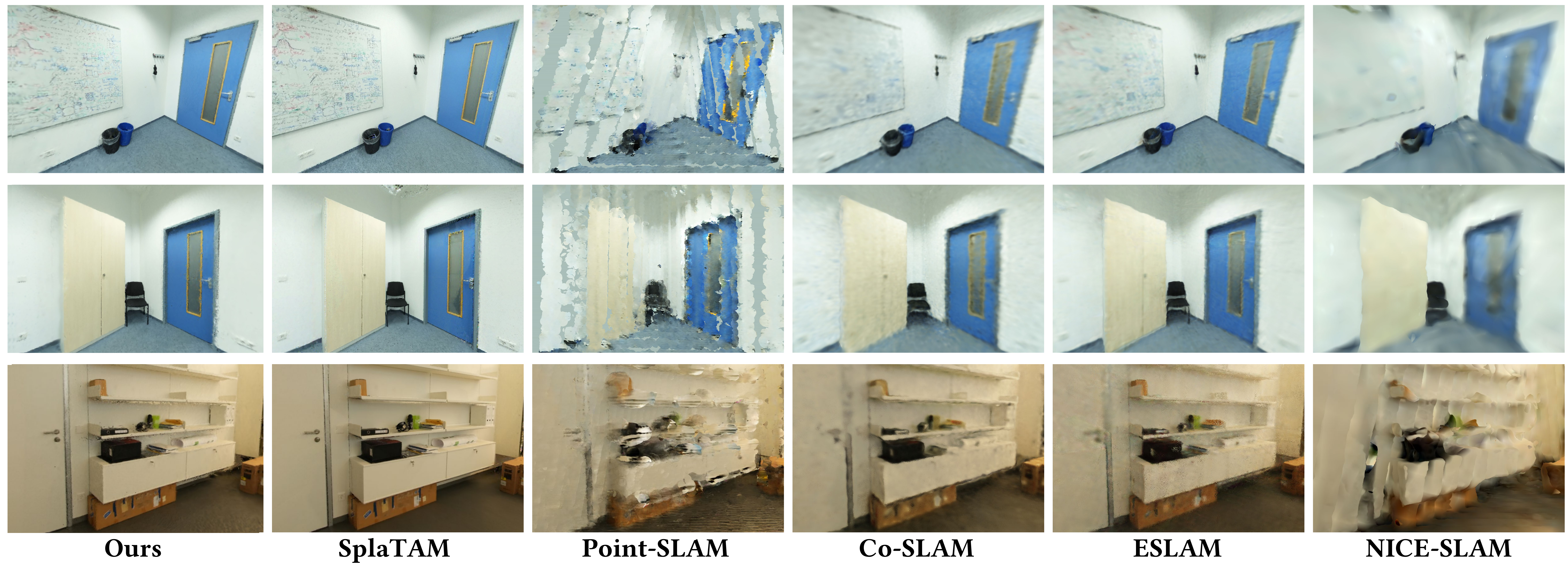}
    \Description[Figure 4: Comparison of novel view synthesis on ScanNet++.] {From left to right: novel vew synthesis of ours, SplaTAM, Point-SLAM, Co-SLAM, ELSAM and NICE-SLAM. The results of Gaussian-based methods (ours and SplaTAM) significantly outperform NeRF-based methods}
    \caption{Comparison of novel view synthesis on ScanNet++.}
    \label{fig:scannetpp_compare}
    \end{figure*}
    \begin{figure*}
    \includegraphics[width=\columnwidth]{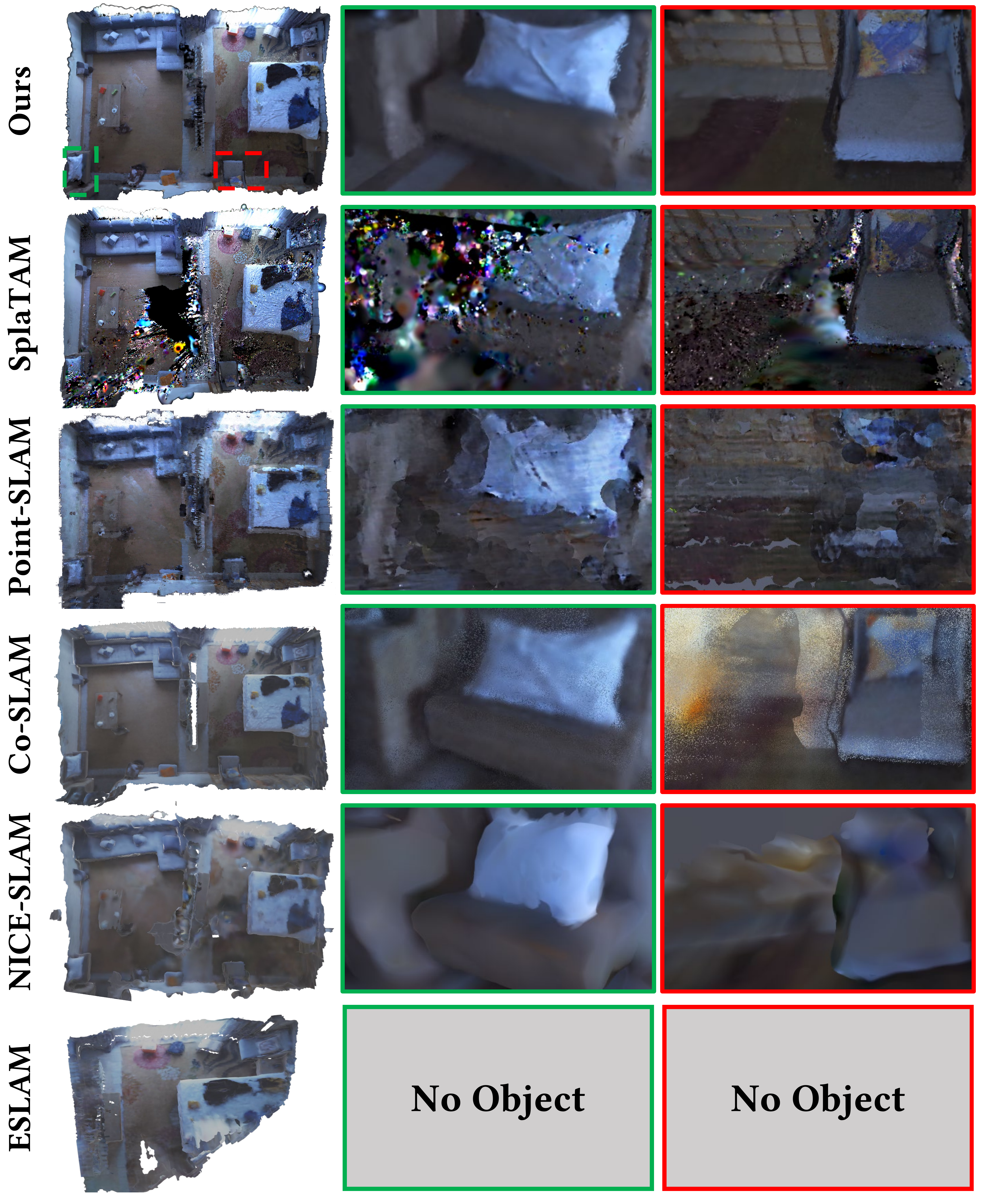}
    \Description[Figure 5: Comparison of reconstruction quality and novel view synthesis in real scanned hotel room scene.]{From top to bottom: top view result and novel view synthesis of ours, SplaTAM, Point-SLAM, Co-SLAM, ELSAM and NICE-SLAM. Our results have better image quality both in the top view and in the novel views}
    \caption{Comparison of reconstruction quality and novel view synthesis in real scanned hotel room scene.}
    \label{fig:ours_dataset_compare}
    \end{figure*}
    \begin{figure*}
        \centering
        \includegraphics[width=\columnwidth]{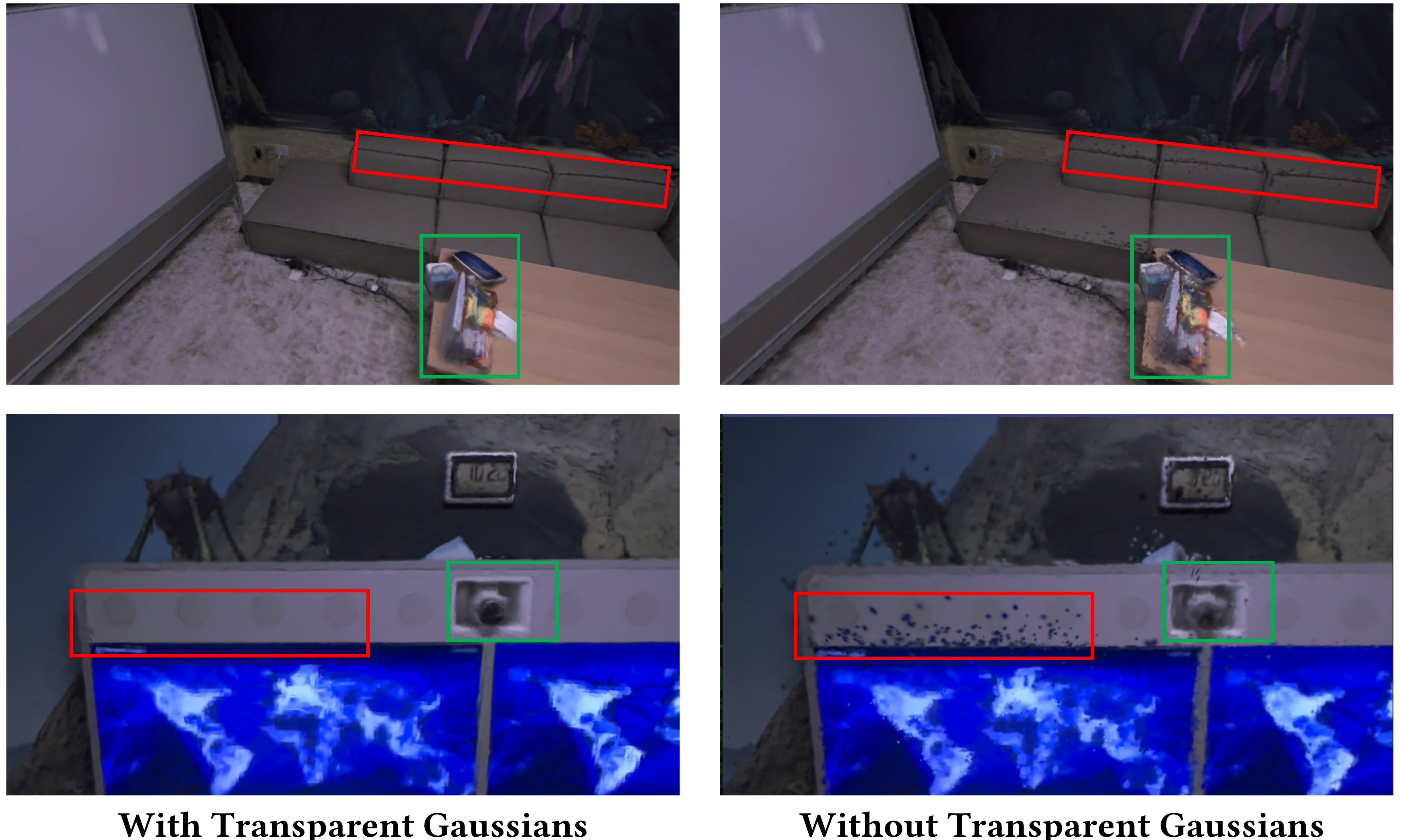}
        \Description[Figure 6: Ablation study on transparent Gaussians]{Left: with transparent Gaussians, Right: Without transparent Gaussians. The scene reconstructed without transparent Gaussians will have noticeable color errors due to occlusion, which can be avoided by transparent Gaussians in our compact Gaussian representation}
        \caption{Ablation study on transparent Gaussians.}
        \label{fig:transparent_ablation}
    \end{figure*}
    \begin{figure*}
        \includegraphics[width=\columnwidth]{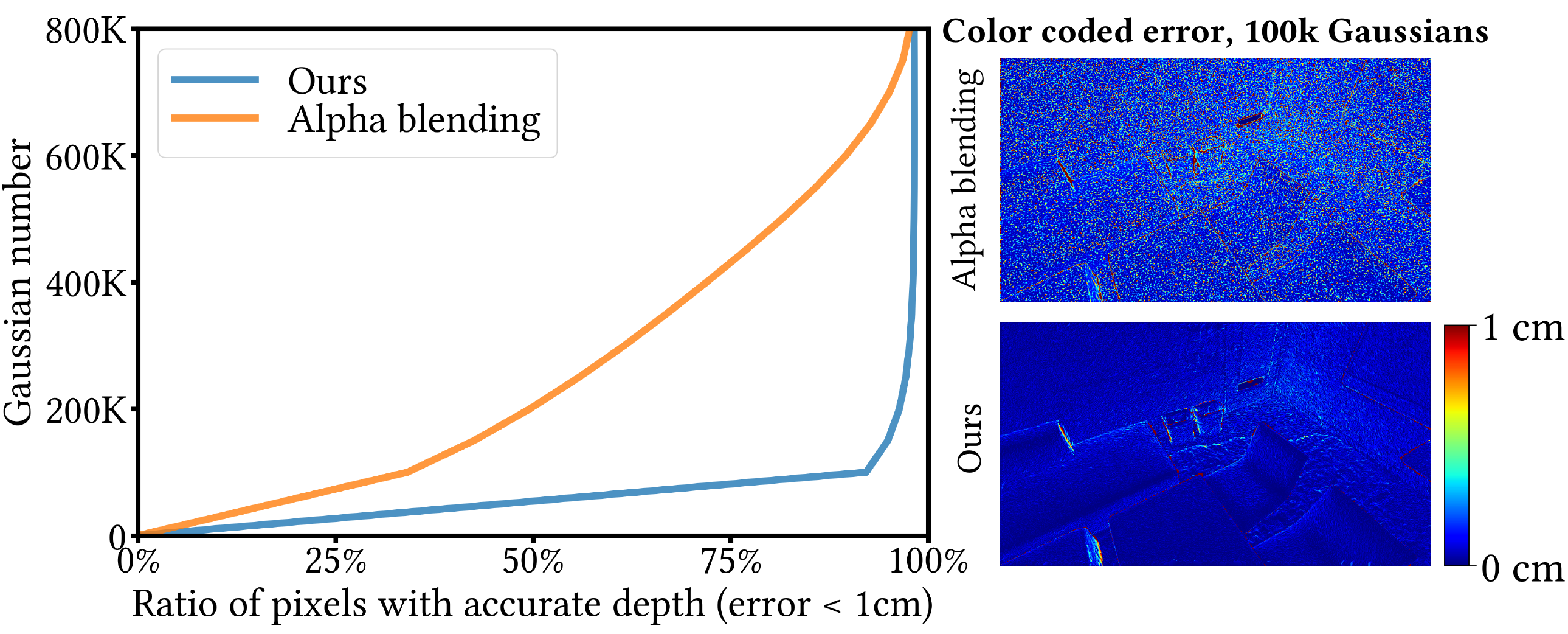}
        \Description[Figure 7: Ablation study on compact Gaussian Representation.]{Left: the plot of Gaussian number versus the ratio of pixels with accurate depth, Right: the color coded depth error of our compact Gaussian representation and alpha blending with 100000 Gaussians. Our compact Gaussian representation can achieve the same depth accuracy with much fewer Gaussians compared with alpha blending}
        \caption{Ablation study on compact Gaussian Representation.}
        \label{fig:compact_ablation}
\end{figure*}}{\begin{figure*}[!htp]
    \begin{minipage}{\textwidth}
    \centering
    \includegraphics[width=\textwidth]{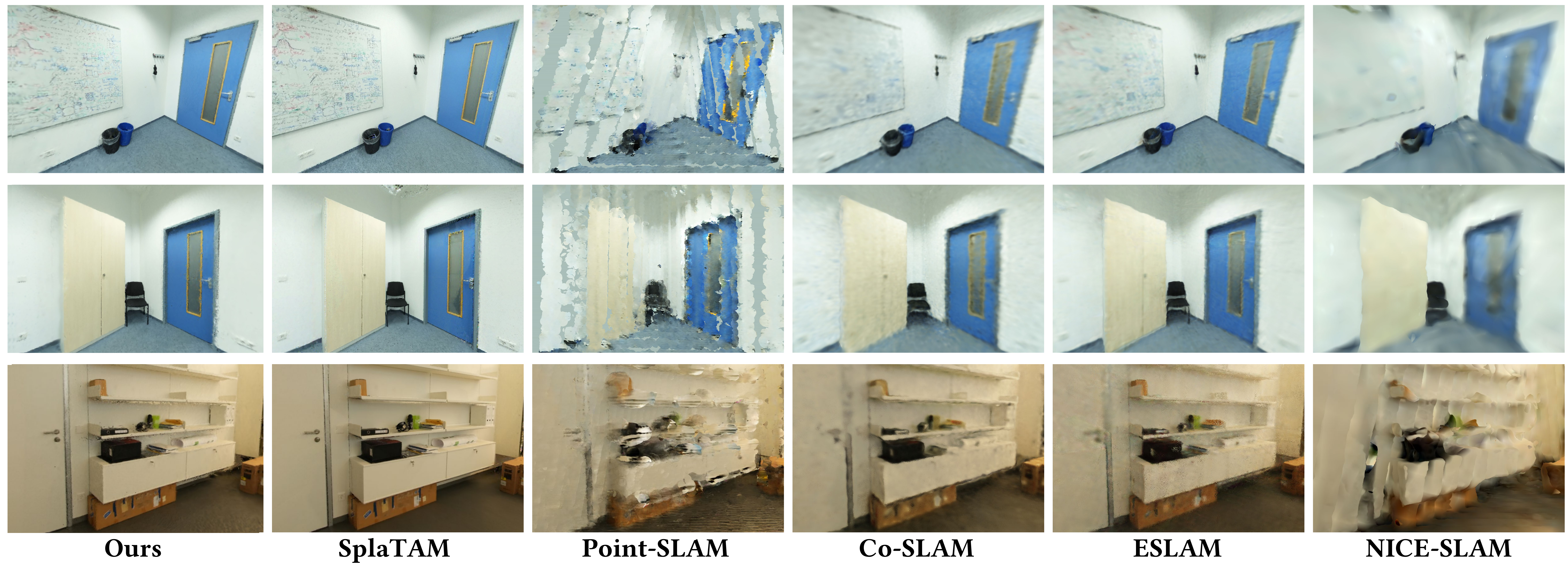}
    \Description[Figure 4: Comparison of novel view synthesis on ScanNet++.] {From left to right: novel vew synthesis of ours, SplaTAM, Point-SLAM, Co-SLAM, ELSAM and NICE-SLAM. The results of Gaussian-based methods (ours and SplaTAM) significantly outperform NeRF-based methods}
    \caption{Comparison of novel view synthesis on ScanNet++.}
    \label{fig:scannetpp_compare}
    
    \end{minipage} \vskip\floatsep
    \begin{minipage}{\columnwidth}
    \includegraphics[width=\columnwidth]{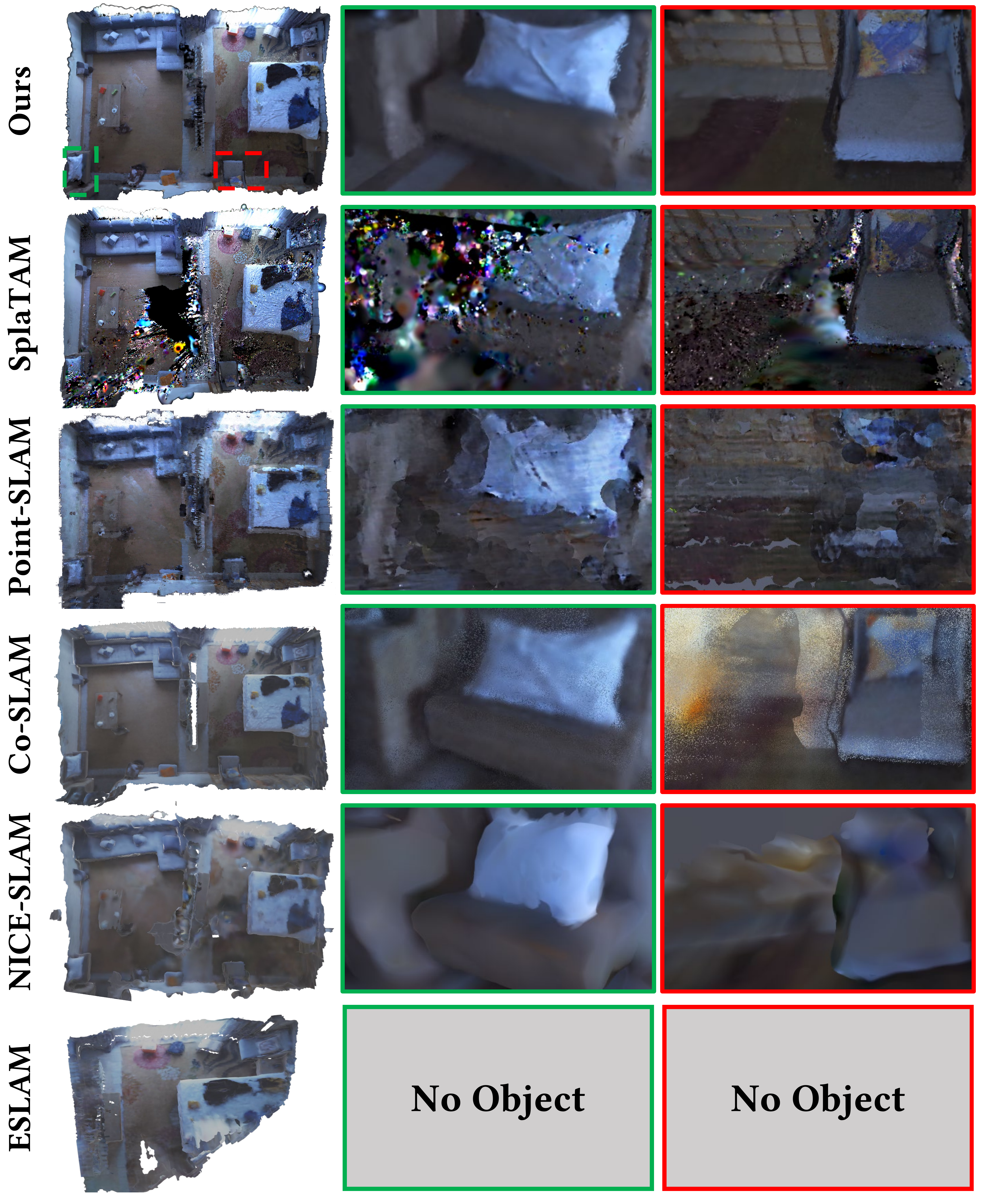}
    \Description[Figure 5: Comparison of reconstruction quality and novel view synthesis in real scanned hotel room scene.]{From top to bottom: top view result and novel view synthesis of ours, SplaTAM, Point-SLAM, Co-SLAM, ELSAM and NICE-SLAM. Our results have better image quality both in the top view and in the novel views}
    \caption{Comparison of reconstruction quality and novel view synthesis in real scanned hotel room scene.}
    \label{fig:ours_dataset_compare}
    
    \end{minipage} \hfill
    \begin{minipage}{\columnwidth}
        \centering
        \includegraphics[width=\columnwidth]{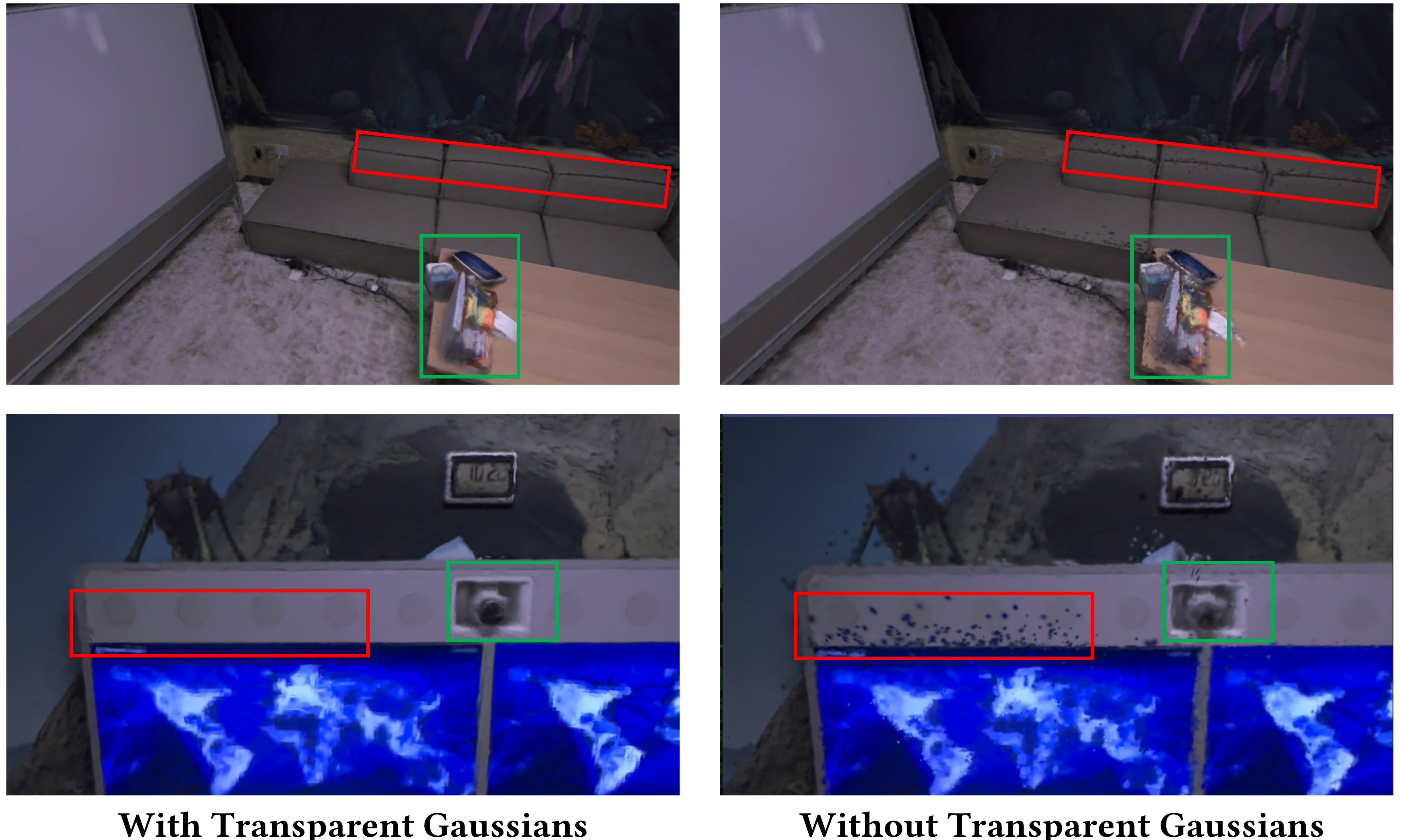}
        \Description[Figure 6: Ablation study on transparent Gaussians]{Left: with transparent Gaussians, Right: Without transparent Gaussians. The scene reconstructed without transparent Gaussians will have noticeable color errors due to occlusion, which can be avoided by transparent Gaussians in our compact Gaussian representation}
        \caption{Ablation study on transparent Gaussians.}
        \label{fig:transparent_ablation}

        \vskip\floatsep
        \includegraphics[width=\columnwidth]{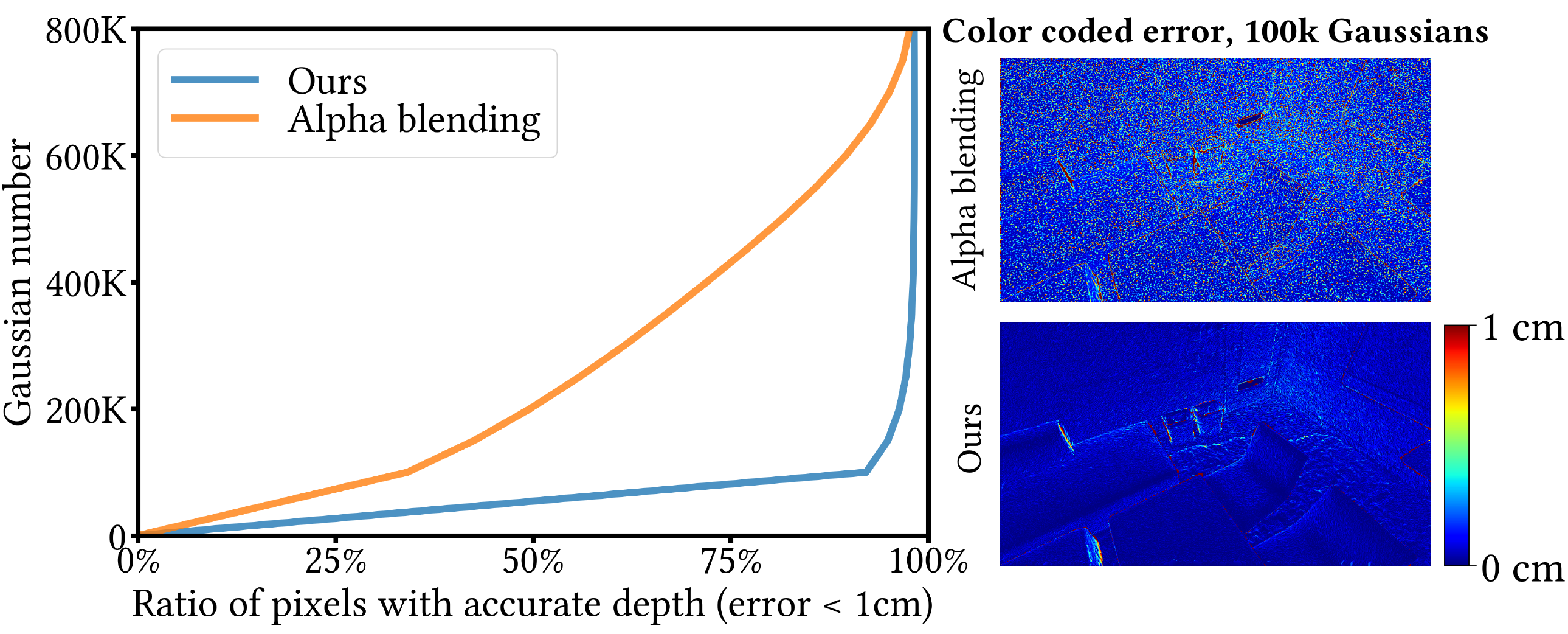}
        \Description[Figure 7: Ablation study on compact Gaussian Representation.]{Left: the plot of Gaussian number versus the ratio of pixels with accurate depth, Right: the color coded depth error of our compact Gaussian representation and alpha blending with 100000 Gaussians. Our compact Gaussian representation can achieve the same depth accuracy with much fewer Gaussians compared with alpha blending}
        \caption{Ablation study on compact Gaussian Representation.}
        \label{fig:compact_ablation}
    \end{minipage}
\end{figure*}}

\begin{figure*}[hp]
    \centering
    \includegraphics[width=\textwidth]{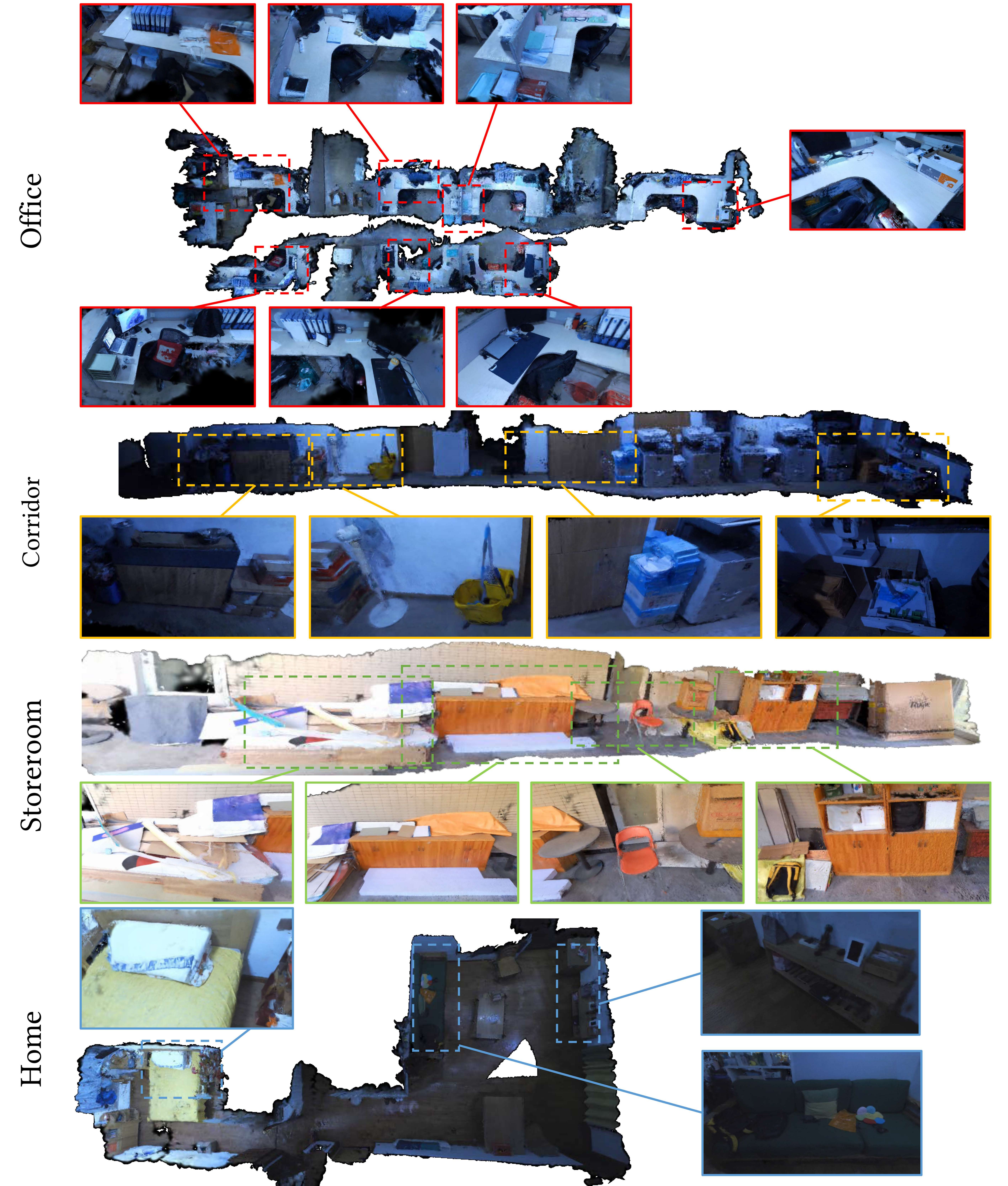}
    \Description[Figure 8: Reconstruction results and novel view synthesis results of real scenes using our system.] {From top to bottom: reconstruction results of office, corridor, storeroom and home. Our method enables large-scale real-world reconstructions in real-time and achieves high-fidelity reconstruction results}
    \caption{Reconstruction results and novel view synthesis results of real scenes using our system.}
    \label{fig:ours_more_res}
\end{figure*}

\clearpage
\appendix
\section*{Supplementary Materials}
\section{Gaussian initialization}
Here we introduce how we compute the covariance matrix $\boldsymbol{\Sigma}_\mathbf{u}$ for each newly added Gaussian for pixel $\mathbf{u}$ in \addmine{detail}. Each new Gaussian $G_{\mathbf{u}}$ is initialized as a flat circle disc. For opaque Gaussians, our goal is to cover the surface as much as possible without largely impacting existing ones. For this, we calculate the distance from the pixel's 3D position $\mathbf{V}_k^g(\mathbf{u})$ to its three nearest Gaussians $G_{1, 2, 3}$ in the scene, and initialize its scale based on the following formula:
\begin{align}
       &\mathbf{s}_{\mathbf{u},1}= \sqrt{\frac{1}{3}\sum_{i=1}^3 \left(||\mathbf{V}_k^g(\mathbf{u}) - \mathbf{p}_i|| - 0.5(a_i+b_i)\right)}, \nonumber \\
       &\mathbf{s}_{\mathbf{u},2}= \mathbf{s}_{\mathbf{u},1} \quad \mathbf{s}_{\mathbf{u},3} = 0.1\mathbf{s}_{\mathbf{u},1}.
\end{align}
Here $a$ is the biggest eigenvalue of the Gaussian covariance matrix and $b$ is the second largest eigenvalue. For the transparent Gaussian, the ratios of its three axes are also set to 1:1:0.1. However, in order to reduce the disruption of the rendering results on other pixels, its maximum scale is limited to 0.01 $m$. 
Finally, the Gaussian orientation $\mathbf{q}_{\mathbf{u}}$ is initialized such that its shortest axis aligns with the pixel's global normal $\mathbf{N}_k^g(\mathbf{u})$. The above initialization method allows the newly added opaque Gaussians to cover the scene surface as much as possible, with little influence on the existing Gaussians in the scene $\mathcal{S}$.

\section{Gaussian optimization}
Since we only optimize the unstable Gaussians, we use $\mathcal{S}_{unstable}$ to render a light transmission map $\hat{\mathbf{T}}——{unstable}$ before optimization, and the loss $L$ is only computed on the pixels:
\begin{equation}
   M_{unstable} = \{\mathbf{u}_{unstable}| \hat{\mathbf{T}}_{unstable}(\mathbf{u}_{unstable}) < 1\}.
\end{equation}

However, in order to achieve fast rendering and optimization, \cite{3DGS} adopts a tile-based rasterizer for Gaussian splatting: the image is divided into $16\times16$ tiles, and each Gaussian is assigned a key that combines view space depth and tile ID and then sorted. In fact, when there are just a few pixels within a tile that need to be rendered, instantiating all Gaussians on this tile is quite inefficient. Therefore, we discard those tiles where the number of pixels that need to be rendered is less than $50\%$. This strategy drastically reduces inefficient computation, and \addmine{increases} the overall speed of the optimization process.

\section{Implementation Details}
We accelerate our system by implementing it in parallel with three threads: one thread for Gaussians optimization, one for front-end online tracking, and the other for back-end graph optimization. The Gaussians optimization and front-end online tracking are implemented by \texttt{python} using the pytorch framework and we write custom \texttt{CUDA} kernels for \addmine{our} rendering process and back propagation. The back-end optimization part is inherited from ORB SLAM2~\cite{ORBSLAM2} and implemented in \texttt{C++}. We also build an interactive viewer using the open-source SIBR~\cite{sibr2020, 3DGS} to visualize the SLAM process and the reconstructed model. In order to achieve free movement and scanning in the scene, we use a laptop with an \texttt{intel i7 10750-H} CPU and \texttt{nvidia 2070 GPU} connected to an Azure Kinect RGBD camera for data acquisition. The RGBD images captured by the camera are transmitted to the desktop computer through a wireless network and the desktop computer completes the SLAM computations, and then the results are sent back to our viewer on the laptop for visualization. 
In all our experiments, we sample 5\% of pixels for the Gaussians adding. For small-scale synthetic dataset Replica~\cite{replica19arxiv}, we set the optimization time window to 6 and optimize 50 iterations. For our large-scale real Azure dataset, we set the optimization time window to 8 and optimize 50 iterations. For TUM-RGBD~\cite{TUM}, we set the optimization time window to 4 and optimize 50 iterations. For ScanNet++~\cite{yeshwanthliu2023scannetpp}, we set the optimization time window to 3 and optimize 75 iterations due to the high-resolution (1752 $\times$ 1168) and sparse viewports. For learning rates, we set $lr_{position}=0.001$, $lr_{SH0}=0.0005$, $lr_{\alpha}=0$, $lr_{\addmine{scale}}=0.004$, $lr_{rotation}=0.001$ on Replica and ScanNet++. And we set $lr_{position}=0.001$, $lr_{SH0}=0.001$, $lr_{\alpha}=0$, $lr_{\addmine{scale}}=\addmine{0.002}$, $lr_{rotation}=0.001$ on our datset and TUM-RGBD. The learning rates of other SH coefficients is 0.05 $\times$ $lr_{SH0}$. \addmine{For the confidence count threshold, we use 
$\delta_{\eta}=100$ for synthetic Replica, $\delta_{\eta}=200$ for TUM-RGBD and $\delta_{\eta}=400$ for large-scale Azure and ScanNet++ scenes.}

\addmine{\section{Dataset Details}
For ScanNet++, we select 4 subsets (8b5caf3398, 39f36da05b, b20a261fdf, f34d532901) for evaluation. The statistics of Azure dataset are listed in Table \ref{tab:statistics_Azure_dataset}. Note that we don't have the ground truth for Azure dataset, so it is mainly used for qualitative demonstration (except the evaluation of time \& memory performance).}
{\small 
    \begin{table}
        \caption{Statistics of Azure dataset} 
        \centering
        \tabcolsep=0.15cm
        \renewcommand\arraystretch{1.2}
        \resizebox{\columnwidth}{!}{

        \begin{tabular}{l c c c c c}
        \toprule[1pt]
            Statistic               &corridor&storeroom &hotel room&home  &office   \\ \hline
            Trajectory Length ($m$) &21.9    &18.9      &39.7      &32.2  &41.0           \\
            Scan Area ($m^2$)       &43.4    &44.3      &56.3       &69.8  &100.2           \\
            Frame Number            &4890    &3310      &4838      &6130  &6889         \\ 
        \bottomrule[1pt]
        \end{tabular}
        }
        \label{tab:statistics_Azure_dataset}
    \end{table} 
}

\section{More comparison results}
\addmine{\subsection{Quantitative results on different datasets}
We add more evaluation on different datasets here. Please note that the low quality 3D model of TUM-RGBD makes it infeasible to evaluate geometry accuracy, as in previous papers. The cameras far apart in ScanNet++ result in tracking failure for classical, NeRF and our SLAM, so only geometry accuracy is evaluated. Our Azure dataset doesn't have the groundtruth camera or 3D model. The comparison of geometry accuracy on Replica is shown in Table \ref{geometry_accuracy_replica}. Similar to the paper, our geometry accuracy still outperforms the other methods except Point-SLAM which doesn't optimize point positions from the perfect depth. SplaTAM and ours degrade in completion because Gaussians cannot complete unscanned regions as NeRF.} The comparison of tracking accuracy on Replica are shown in Table \ref{tab:track_acc_replica}. Our method consistently outperforms the NeRF SLAM and the concurrent Gaussian SLAM method. We believe this is due to our back-end graph optimization based on ORB landmarks. Please note that in order to ensure fairness in comparison, although there is no noise in the depth images on Replica, we still used frame-to-model ICP, just without applying the bilateral filter to the depth input. \addmine{We also report the time and memory performance on TUM-RGBD in Table \ref{tab:time_memory_performance_TUMRGBD}. Our system achieves the highest scanning speed and lowest memory cost.}
{\small 
    \begin{table}
        \caption{Comparison of geometry accuracy on Replica.} 
        \centering
        \tabcolsep=0.15cm
        \renewcommand\arraystretch{1.2}
        \resizebox{\columnwidth}{!}{

        \begin{tabular}{c c c c c }
        \toprule[1pt]
            Method                            & Acc.$\downarrow$    & Acc. Ratio$\uparrow$   & Comp.$\downarrow$     & Comp. Ratio$\uparrow$ \\ \hline
            {NICE-SLAM\shortcite{nice_slam}}  & 2.84                & 84.44                 & 2.31                   &  84.97    \\ 
            {Co-SLAM\shortcite{co_slam}}      & 2.33                & 88.89                 & \underline{1.63}       &  \underline{89.94}    \\ 
            {ESLAM\shortcite{eslam}}          & 1.47                & 91.44                 & \textbf{1.11}          &  \textbf{93.84}    \\ 
            {Point-SLAM\shortcite{point_slam}} & \textbf{0.61}      & \textbf{99.94}        & 2.42                   &  86.85   \\ 
            {SplaTAM\shortcite{splatam}}       & 2.88               & 73.89                 & 3.57                   &  71.68    \\ 
            {Ours}                             & \underline{0.75}   & \underline{98.87}     & 2.81                   &  82.76     \\
        \bottomrule[1pt]
        \end{tabular}
        }
        \label{geometry_accuracy_replica}
    \end{table} 
}

{\small 
   \begin{table}
       \caption{Comparison of tracking accuracy (unit: $cm$) on Replica.}
       \centering
       \tabcolsep=0.15cm
       \renewcommand\arraystretch{1.2}
       \resizebox{\columnwidth}{!}{
       \begin{tabular}{c c c c c c c c c c}
       \toprule[1pt]
           Method     & Rm 0  & Rm 1 & Rm 2 & Off 0& Off 1 & Off 2 & Off 3 & Off 4 & Avg. \\ \hline
           NICE-SLAM\shortcite{nice_slam}  & 0.97  & 1.31 & 1.07 & 0.88 & 1.00  & 1.06  & 1.10  & 1.13  & 1.06 \\
           Co-SLAM\shortcite{co_slam}    & 0.69  & 0.59 & 0.73 & 0.87 & 0.47  & 2.16  & 1.30  & 0.62  & 0.93 \\ 
           ESLAM\shortcite{eslam}     & 0.66  & 0.62 & 0.55 & 0.44	& 0.43	& 0.50	& 0.66	& \underline{0.53}  & 0.55  \\
           Point-SLAM\shortcite{point_slam} & 0.54  & 0.43 & 0.34 & \underline{0.36}	& 0.45	& 0.44	& 0.63	& 0.72  & 0.49  \\
           SplaTAM\shortcite{splatam}    & \underline{0.47}  & \underline{0.42} & \underline{0.32} & 0.46	& \underline{0.24}  & \underline{0.28}  & \underline{0.39}  & 0.56  & \underline{0.39}   \\
           Ours       & \textbf{0.20}  & \textbf{0.18} & \textbf{0.13} & \textbf{0.22} & \textbf{0.12}  & \textbf{0.22}  & \textbf{0.20}	& \textbf{0.19}  & \textbf{0.18}  \\
       \bottomrule[1pt]
       \end{tabular}
       }
       \label{tab:track_acc_replica}
   \end{table} 
}


Finally we compare the rendering quality. The results of training view synthesis quality on Replica are reported in Table \ref{tab:rendering_training_replica}. Please note that this comparison is actually unfair as Point-SLAM~\shortcite{point_slam} takes the ground-truth depth maps as input to help sampling the 3D volume for rendering. In contrast, our method and SplaTAM~\shortcite{splatam} do not require any auxiliary input. Even so, our method still achieves a rendering quality comparable with Point-SLAM and SplaTAM, and consistently outperforms the other NeRF SLAM methods. \addmine{
The quantitative comparison of novel-view synthesis on ScanNet++ testing views is reported in Table \ref{tab:novel_view_synthesis}. Our method is comparable to SplaTAM and outperforms the other NeRF-SLAM methods.}
{\small 
    \begin{table}
        \caption{Comparison of time and memory performance on TUM-RGBD.} 
        \centering
        \tabcolsep=0.15cm
        \renewcommand\arraystretch{1.2}

        \begin{tabular}{c c c c}
        \toprule[1pt]
            Method         & FPS$\uparrow$     & Memory (MB)$\downarrow$   \\ \hline
            NICE-SLAM\shortcite{nice_slam}        & 0.06     & \underline{9930}           \\
            CO-SLAM\shortcite{co_slam}            & \underline{7.18}    & 18607         \\ 
            ESLAM\shortcite{eslam}                & 0.30    & 18617           \\
            Point-SLAM\shortcite{point_slam}     & 0.26    & 11000           \\
            SplaTAM\shortcite{splatam}           & 0.14    & 12100          \\
            Ours                                 & \textbf{21.74}    & \textbf{3563}           \\

        \bottomrule[1pt]
        \end{tabular}
        \label{tab:time_memory_performance_TUMRGBD}
    \end{table} 
}

{\small 
    \begin{table}
            \caption{Comparison of train view synthesis on Replica.}
        \centering
        \tabcolsep=0.15cm
        \renewcommand\arraystretch{1.2}
        \resizebox{\columnwidth}{!}{
        \begin{tabular}{c l c c c c c c c c c}
        \toprule[1pt]
            Method     & Metric               & Rm 0   & Rm 1  & Rm 2  & Off 0 & Off 1 & Off 2 & Off 3 & Off 4 & Avg. \\ \hline
            \multirow{3}{*}{\makecell{NICE-SLAM\\\shortcite{nice_slam}}} 
                        &PSNR$\uparrow$      & 24.72  & 26.79 & 27.06 & 30.21 & 32.78 & 26.59 & 26.22 & 24.74 & 27.39 \\
                        &SSIM$\uparrow$      & 0.787   & 0.799  & 0.807  & 0.881  & 0.906  & 0.816  & 0.801  & 0.834  & 0.829 \\
                        &LPIPS$\downarrow$   & 0.431   & 0.372  & 0.329  & 0.322  & 0.275  & 0.321  & 0.288  & 0.333 & 0.334 \\ \hline
            \multirow{3}{*}{\makecell{Co-SLAM\\\shortcite{co_slam}}} 
                        &PSNR$\uparrow$      & 28.88  & 28.51 & 29.37 & 35.44 & 34.63 & 26.56 & 28.79 & 32.16 & 30.54 \\
                        &SSIM$\uparrow$      &  0.892  & 0.843  & 0.851  &  0.854 &  0.826 & 0.814  &  0.866 & 0.856  & 0.850  \\
                        &LPIPS$\downarrow$   &  0.213  &  0.205 & 0.215 & 0.177  & 0.181  & 0.172  & 0.163  & 0.176   & 0.188 \\ \hline
            \multirow{3}{*}{\makecell{ESLAM\\\shortcite{eslam}}} 
                        &PSNR$\uparrow$      & 26.96  & 28.98 & 29.80 & 35.04 & 33.81 & 30.08 & 30.01 & 31.34 & 30.75 \\
                        &SSIM$\uparrow$      &  0.821  & 0.837  & 0.843  & 0.902  & 0.873  & 0.865 & 0.881  & 0.886 & 0.863 \\
                        &LPIPS$\downarrow$   &  0.171  & 0.173  & 0.187  & 0.172  & 0.181  &  0.186 & 0.172  & 0.174 & 0.177 \\ \hline
            \multirow{3}{*}{\makecell{Point-SLAM\\\shortcite{point_slam}}} 
                        &PSNR$\uparrow$      & \textbf{32.40}  & \underline{34.08} &  \underline{35.50} &  \underline{38.26} &  \underline{39.16} & \textbf{33.99} & \textbf{33.48} &  \underline{33.49} &  \underline{35.17} \\
                        &SSIM$\uparrow$      &  \textbf{0.974}     & \underline{0.977} & \underline{0.982} & \underline{0.983} & \underline{0.986}  & 0.960  & \underline{0.960}  & \underline{0.979}  & \underline{0.975}  \\
                        &LPIPS$\downarrow$   &  \underline{0.113}    & 0.116  & \underline{0.111} & 0.100 & 0.118  & 0.156  &  0.132 &  \underline{0.142} & 0.124 \\ \hline
            \multirow{3}{*}{\makecell{SplaTAM\\\shortcite{splatam}}} 
                        &PSNR$\uparrow$      & \underline{32.31}  & 33.36 & 34.78 & 38.16 & 38.49 & 31.66 & 29.24 & 31.54 & 33.69 \\
                        &SSIM$\uparrow$      &  \textbf{0.974}     & 0.966  & \textbf{0.983}  & 0.982  & 0.980  & \underline{0.962}  & 0.948  & 0.946  & 0.968  \\
                        &LPIPS$\downarrow$   &  \textbf{0.072}     & \textbf{0.101}  & \textbf{0.073}  & \underline{0.084}  & \underline{0.095}  & \textbf{0.102}  & \textbf{0.123}  & 0.157  & \textbf{0.101} \\ \hline
            \multirow{3}{*}{\makecell{Ours}} 
                        &PSNR$\uparrow$      & 31.56  & \textbf{34.21} & \textbf{35.57} & \textbf{39.11} & \textbf{40.27} & \underline{33.54}  & \underline{32.76} & \textbf{36.48} & \textbf{35.43} \\
                        &SSIM$\uparrow$      & 0.967      & \textbf{0.979}  & 0.981  & \textbf{0.990}   &  \textbf{0.992}  & \textbf{0.981}  &  \textbf{0.981} &  \textbf{0.984} & \textbf{0.982} \\
                        &LPIPS$\downarrow$   & 0.131  & \underline{0.105} & 0.115 & \textbf{0.068}  & \textbf{0.075} & \underline{0.134}  & \underline{0.128} & \textbf{0.117} & \underline{0.109} \\
        \bottomrule[1pt]
        \end{tabular}
        }
        \label{tab:rendering_training_replica}
    \end{table} 
}
\begin{figure*}[h]
    \centering
    \includegraphics[width=\textwidth]{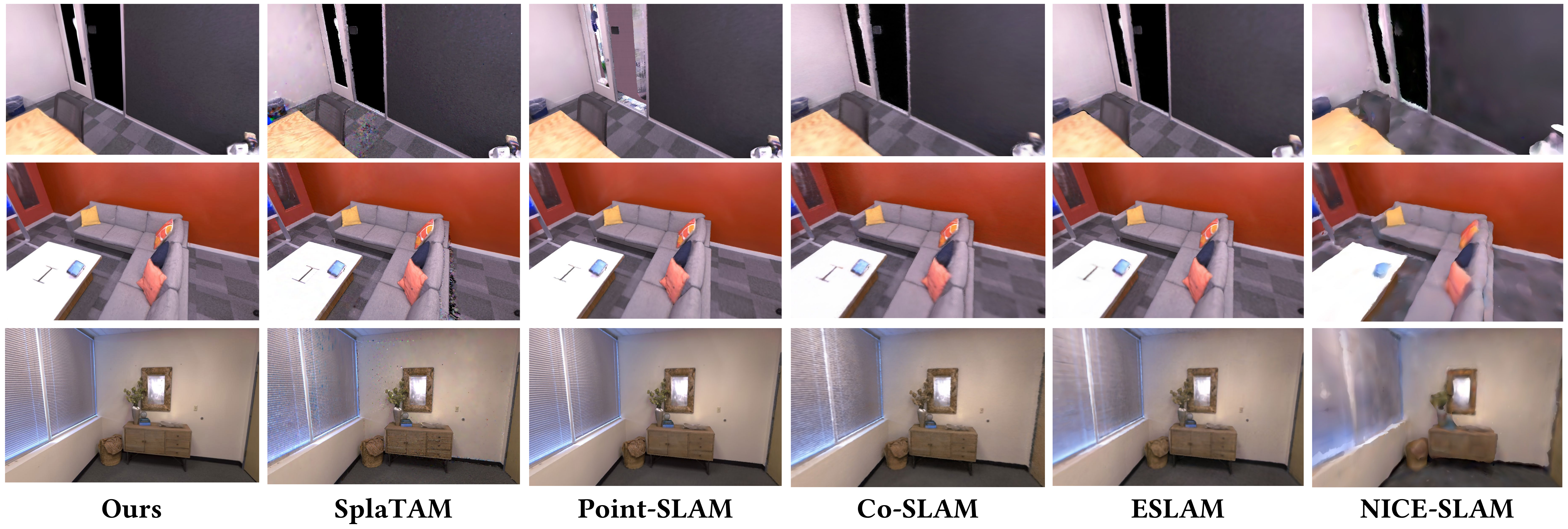}
    \caption{Comparison of novel view synthesis on Replica.}
    \label{fig:replica_compare}
\end{figure*}

{\small 
    \begin{table}
            \caption{Comparison of novel view synthesis on ScanNet++.} 
        \centering
        \tabcolsep=0.15cm
        \renewcommand\arraystretch{1.2}

        \begin{tabular}{c c c c}
        \toprule[1pt]
            Method                               & PSNR$\uparrow$     & SSIM$\uparrow$    & LPIPS$\downarrow$   \\ \hline
            NICE-SLAM\shortcite{nice_slam}       & 23.71    &0.797     &0.341      \\
            CO-SLAM\shortcite{co_slam}           &23.20     &0.837     &0.413    \\ 
            ESLAM\shortcite{eslam}               & 27.06    &0.856     &0.322      \\
            Point-SLAM\shortcite{point_slam}     & 21.85    &0.802     &0.404      \\
            SplaTAM\shortcite{splatam}           & \textbf{27.77}    &\underline{0.864}     &\textbf{0.233}      \\
            Ours                                 & \underline{27.27}  &\textbf{0.872}        &\underline{0.295}      \\

        \bottomrule[1pt]
        \end{tabular}
        \label{tab:novel_view_synthesis}
    \end{table} 
}

\addmine{\subsection{Comparison with more SLAM systems} 
We show more comparisons with ElasticFusion~\cite{elasticfusion}, BundleFusion~\cite{bundlefusion}, and GO-SLAM~\cite{goslam}. The tracking accuracy is evaluated on Replica and TUM-RGBD and the results are listed in Table \ref{tab:tracking_comparison_with_classical_slam}. we also evaluate the geometric accuracy on Replica as shown in Table \ref{tab:geometry_comparison_with_classical_slam}. Our method achieves comparable geometry accuracy as BundleFusion. GO-SLAM's completion numbers are worse than those of its paper due to considering unscanned regions for fair comparison.}
{\small 
    \begin{table}
            \caption{Comparison of tracking accuracy (unit: $cm$) with more SLAM systems.} 
        \centering
        \tabcolsep=0.15cm
        \renewcommand\arraystretch{1.2}
        \begin{tabular}{c c c c}
        \toprule[1pt]
            Method                                 & Replica               & TUM-RGBD   \\ \hline
            ElasticFusion\shortcite{elasticfusion} & 1.13                  & 2.07       \\ 
            BundleFusion\shortcite{bundlefusion}   & 0.46                  & 1.63       \\
            GO-SLAM\shortcite{goslam}              & \underline{0.37}      & \underline{1.28}       \\
            Ours                                   & \textbf{0.18 }        & \textbf{1.06}       \\
        \bottomrule[1pt]
        \end{tabular}
        \label{tab:tracking_comparison_with_classical_slam}
    \end{table} 
}
{\small 
    \begin{table}
        \caption{Comparison of geometry accuracy on Replica with more SLAM systems.} 
        \centering
        \tabcolsep=0.15cm
        \renewcommand\arraystretch{1.2}
        \resizebox{\columnwidth}{!}{

        \begin{tabular}{c c c c c }
        \toprule[1pt]
            Method                                   & Acc.$\downarrow$    & Acc. Ratio$\uparrow$   & Comp.$\downarrow$     & Comp. Ratio$\uparrow$ \\ \hline
            {ElasticFusion\shortcite{elasticfusion}} & 1.13                & 96.33                  & \underline{4.43}                  &  75.25    \\ 
            {BundleFusion\shortcite{bundlefusion}}   & \underline{0.77}    & \textbf{99.88}         & 5.35                  &  \underline{76.69}    \\
            {GO-SLAM\shortcite{goslam}}              & 2.51                & 76.93                  & 5.11                  &  65.10    \\ 
            {Ours}                                   & \textbf{0.75}       & \underline{98.87}      & \textbf{2.81}                  &  \textbf{82.76}    \\ 
        \bottomrule[1pt]
        \end{tabular}
        }
        \label{tab:geometry_comparison_with_classical_slam}
    \end{table} 
}
\section{More ablation studies}
\subsection{\addmine{Ablation} study on sampled pixel number} 
Here we \addmine{evaluate} the influence of the number of sampled pixels for adding Gaussians. We set the sampling ratio to 5\%, 10\%, and 20\% for each frame to reconstruct Replica office0 and reported the image quality metrics. The results suggest that even if we sample a small number of images for reconstruction, the image quality is not significantly affected.
{\small 
    \begin{table}
            \caption{\addmine{Ablation} study on sampled pixel number.} 
        \centering
        \tabcolsep=0.15cm
        \renewcommand\arraystretch{1.2}

        \begin{tabular}{c c c c}
        \toprule[1pt]
            Sample ratio  & PSNR$\uparrow$     & SSIM$\uparrow$    & LPIPS$\downarrow$   \\ \hline
            $5\%$         & 39.01              & 0.965  &  0.072    \\ 
            $10\%$        & \textbf{39.63}     & \underline{0.970}  &  \underline{0.051}      \\
            $20\%$        & \underline{39.31}  & \textbf{0.972}  &  \textbf{0.044}      \\
        \bottomrule[1pt]
        \end{tabular}
        \label{tab:gaussian_num_ablation1}
    \end{table} 
}

\subsection{\addmine{Ablation} study on stable/unstable Gaussians} 
We test the impact of our proposed stable/unstable Gaussians on time performance. We report the optimization time per iteration, for optimizing all Gaussians using the whole image, optimizing only unstable Gaussians using the whole image, and optimizing only unstable Gaussians using only the pixels covered by them. As seen in Table \ref{tab:stable_gaussian_ablation}, our strategy greatly improves the optimization speed.
{\small 
    \begin{table}
            \caption{\addmine{Ablation} study on stable/unstable Gaussians.} 
        \centering
        \tabcolsep=0.15cm
        \renewcommand\arraystretch{1.2}
        \resizebox{\columnwidth}{!}{

        \begin{tabular}{l c c c}
        \toprule[1pt]
            Scene                                    & Storeroom       & Hotel room           & Home   \\ \hline
            $\mathcal{S}$                            & 9.8ms  & 8.4ms  &  8.9ms     \\ 
            $\mathcal{S}_{unstable}$, all pixels     & \underline{7.4ms}  & \underline{6.5ms}  &  \underline{6.4ms}      \\
            $\mathcal{S}_{unstable}$, unstable pixels& \textbf{5.2ms}  & \textbf{4.7ms}  &  \textbf{4.3ms}      \\
        \bottomrule[1pt]
        \end{tabular}
        }
        \label{tab:stable_gaussian_ablation}
    \end{table} 
}

\addmine{\subsection{Ablation study on depth rendering}
Our depth blending is tightly coupled with our Gaussian adding and state management, so in the paper we show that alpha blending yields much more Gaussians through the comparison with SplaTAM which uses alpha blending (987524 vs 7155880). Here for better ablation study, we first use our depth blending to determine the adding of opaque/transparent Gaussians as well as the states, and then replace our depth blending with alpha blending for optimization. The results are listed in Table \ref{tab:depth_rendering_ablation}. We can see with similar Gaussian numbers, our depth blending outperforms alpha blending in terms of geometry accuracy and tracking accuracy.}

{\small 
    \begin{table}
            \caption{Ablation study on depth rendering.} 
        \centering
        \tabcolsep=0.15cm
        \renewcommand\arraystretch{1.2}
        \resizebox{\columnwidth}{!}{

        \begin{tabular}{c c c c c c c}
        \toprule[1pt]
            Method          & Acc.$\downarrow$    & Acc. Ratio$\uparrow$   & Comp.$\downarrow$     & Comp. Ratio$\uparrow$ &ATE (cm)$\downarrow$ & Gaussian Number \\ \hline
            {Alpha-blending}& 2.48                & 70.54                  & 3.32                  &  75.01    & 1.24   & 468916 \\ 
            {Ours}          & \textbf{0.75}       & \textbf{98.87}         & \textbf{2.81}         &  \textbf{82.76} & \textbf{0.18}   & \textbf{431692} \\ 
        \bottomrule[1pt]
        \end{tabular}
        }
        \label{tab:depth_rendering_ablation}
    \end{table} 
}

\addmine{\subsection{Ablation study on confidence count threshold}
The ablation study on confidence count threshold $\delta_{\eta}$ on Replica is shown in Table \ref{tab:confidence_count_threshold_ablation}. As $\delta_{\eta}$ increases, the Gaussians will be in the unstable state for a longer time, resulting in a slower speed. On the other hand, if $\delta_{\eta}$ is small, the Gaussians may not be fully optimized, yielding reduced rendering quality.}
{\small 
    \begin{table}
            \caption{Ablation study on confidence count threshold.} 
        \centering
        \tabcolsep=0.15cm
        \renewcommand\arraystretch{1.2}
        \begin{tabular}{c c c c}
        \toprule[1pt]
            $\delta_{\eta}$     & PSNR$\uparrow$     & FPS$\uparrow$  \\ \hline
            50                  & 33.63              & \textbf{18.21} \\ 
            100                 & \textbf{35.43}     & \underline{17.31} \\
            200                 & 35.12              & 16.59  \\
            400                 & \underline{35.37}  & 15.49  \\
        \bottomrule[1pt]
        \end{tabular}
        \label{tab:confidence_count_threshold_ablation}
    \end{table} 
}

\addmine{\subsection{Ablation study on backend pose optimization}
Here we evaluate the effect of backend pose optimization. We test the tracking accuracy on Replica and TUM-RGBD, and report the results in Table \ref{tab:backend_pose_optimization_ablation_tracking}. We also test the geometry accuracy on Replica and report the results in Table \ref{tab:backend_pose_optimization_ablation_geometry}. On high-quality images on Replica, we achieve relatively high accuracy using only the frontend ICP. However, on low-quality images on TUM-RGBD, we rely more on the ORB backend, because the ICP may be performed on the Gaussians still under optimization.}
{\small 
    \begin{table}
            \caption{Ablation study on backend pose optimization in terms of tracking accuracy .} 
        \centering
        \tabcolsep=0.15cm
        \renewcommand\arraystretch{1.2}
        \resizebox{\columnwidth}{!}{

        \begin{tabular}{l c c c}
        \toprule[1pt]
            Method                             & Replica      & \makecell{TUM-RGBD\\fr1\_desk} & \makecell{TUM-RGBD\\fr2\_xyz} \\ \hline
            ElasticFusion without backend      & 0.68             & 2.93              & 1.32      \\
            ElasticFusion                      & 1.13             & \underline{2.53}  & \underline{1.17}      \\ 
            Ours without backend               & \underline{0.22} & 5.39              & 1.63       \\
            Ours                               & \textbf{0.18}    & \textbf{1.66}     & \textbf{0.38}       \\
        \bottomrule[1pt]
        \end{tabular}
        }
        \label{tab:backend_pose_optimization_ablation_tracking}
    \end{table} 
}

{\small 
    \begin{table}
            \caption{Ablation study on backend pose optimization in terms of geometry accuracy .} 
        \centering
        \tabcolsep=0.15cm
        \renewcommand\arraystretch{1.2}
        \resizebox{\columnwidth}{!}{

        \begin{tabular}{l c c c c }
        \toprule[1pt]
            Method                        & Acc.$\downarrow$    & Acc. Ratio$\uparrow$   & Comp.$\downarrow$ & Comp. Ratio$\uparrow$ \\ \hline
            ElasticFusion without backend & 1.05                & \underline{98.32}      & 4.33              &  77.69    \\ 
            ElasticFusion                 & 1.13                & 96.33                  & 4.43              &  75.25    \\
            Ours without backend          & \underline{0.95}    & 97.57                  & \textbf{2.74}     &  \textbf{83.17}\\ 
            Ours                          & \textbf{0.75}       & \textbf{98.87}         & \underline{2.81}  &  \underline{82.76}    \\ 
        \bottomrule[1pt]
        \end{tabular}
        }
        \label{tab:backend_pose_optimization_ablation_geometry}
    \end{table} 
}

\addmine{\subsection{Ablation study on SH coefficients}
Here we report the rendering quality using SHs and RGB colors. We report the results of training view synthesis on Replica in Table \ref{tab:SH_coefficients_ablation_train}. We also report the results of novel view synthesis on ScanNet++ in Table \ref{tab:SH_coefficients_ablation_novel}. We can notice that using SH coefficients has better rendering quality.}
{\small 
    \begin{table}
            \caption{Ablation study on SH coefficients in terms of training view synthesis.} 
        \centering
        \tabcolsep=0.15cm
        \renewcommand\arraystretch{1.2}

        \begin{tabular}{c c c c}
        \toprule[1pt]
            Color     & PSNR$\uparrow$     & SSIM$\uparrow$    & LPIPS$\downarrow$   \\ \hline
            RGB       & 33.90     & 0.951  &  0.113    \\ 
            SH        & \textbf{35.43}     & \textbf{0.982}  &  \textbf{0.109}      \\
        \bottomrule[1pt]
        \end{tabular}
        \label{tab:SH_coefficients_ablation_train}
    \end{table} 
}
{\small 
    \begin{table}
            \caption{Ablation study on SH coefficients in terms of novel view synthesis .} 
        \centering
        \tabcolsep=0.15cm
        \renewcommand\arraystretch{1.2}

        \begin{tabular}{c c c c}
        \toprule[1pt]
            Color     & PSNR$\uparrow$     & SSIM$\uparrow$    & LPIPS$\downarrow$   \\ \hline
            RGB       & 25.67              &0.855              & 0.304    \\ 
            SH        & \textbf{27.27}     & \textbf{0.874}    & \textbf{0.291}      \\
        \bottomrule[1pt]
        \end{tabular}
        \label{tab:SH_coefficients_ablation_novel}
    \end{table} 
}
\addmine{\subsection{Ablation study on outlier pruning}
Here we evaluate the influence of our outlier pruning strategy. We report the number of Gaussians every 1000 frames in the Azure hotel room scene in Table \ref{tab:outlier_pruning_ablation} and the results show that the Gaussian number will increase significantly without outlier pruning.} 
{\small 
    \begin{table}
            \caption{Ablation study on outlier pruning.} 
        \centering
        \tabcolsep=0.15cm
        \renewcommand\arraystretch{1.2}

        \begin{tabular}{c c c c}
        \toprule[1pt]
            Frame ID     & Ours               & Ours without outlier pruning \\ \hline
            1000\#       & \textbf{151947}    &211390    \\ 
            2000\#       & \textbf{268625}    &382736    \\ 
            3000\#       & \textbf{507987}    &797242    \\ 
            4000\#       & \textbf{675818}    &1073366    \\ 
        \bottomrule[1pt]
        \end{tabular}
        \label{tab:outlier_pruning_ablation}
    \end{table} 
}

\end{document}